%% file: main.tex
\documentclass[10pt,twocolumn,letterpaper]{article}

\usepackage[pagenumbers]{cvpr}

\input{preamble}

\definecolor{cvprblue}{rgb}{0.21,0.49,0.74}
\usepackage[pagebackref,breaklinks,colorlinks,allcolors=cvprblue]{hyperref}
\hypersetup{
  pdftitle={A Mesh to Remember: Persistent Appearance for Interactive Video Worlds},
  pdfauthor={Joonghyuk Shin, Alex Trevithick, Katarína Tóthová, Liyuan Zhu, Shengyu Huang, Jaesik Park, Shalini De Mello, Laura Leal-Taixé, Riccardo de Lutio, Haithem Turki},
}

\def\confName{CVPR}
\def\confYear{2027}

\title{A Mesh to Remember:\\
Persistent Appearance for Interactive Video Worlds}

\author{%
Joonghyuk Shin\textsuperscript{1,2*}\quad
Alex Trevithick\textsuperscript{1}\quad
Katar\'{\i}na T\'othov\'a\textsuperscript{1}\quad
Liyuan Zhu\textsuperscript{1,3*}\quad
Shengyu Huang\textsuperscript{1}\\
Jaesik Park\textsuperscript{2}\quad
Shalini De Mello\textsuperscript{1}\quad
Laura Leal-Taix\'e\textsuperscript{1}\quad
Riccardo de Lutio\textsuperscript{1\dag}\quad
Haithem Turki\textsuperscript{1\dag}\\[4pt]
\textsuperscript{1}NVIDIA\qquad
\textsuperscript{2}Seoul National University\qquad
\textsuperscript{3}Stanford University\\[2pt]
{\normalsize\textsuperscript{*}Work done during an internship at NVIDIA\qquad\textsuperscript{\dag}Equal supervision}\\[8pt]
{\normalsize\url{https://research.nvidia.com/labs/sil/projects/m2r/}}%
}
\begin{document}
\twocolumn[{%
\renewcommand\twocolumn[1][]{#1}%
\maketitle
\vspace{-24pt}
\input{sec/0_teaser}}
]
\input{sec/0_abstract}    
\input{sec/1_intro}
\input{sec/2_related}
\input{sec/3_method}

\input{sec/4_experiments}

\input{sec/5_conclusion}
{
    \small
    \FloatBarrier
    \bibliographystyle{ieeenat_fullname}
    \bibliography{main}
}
\input{sec/X_suppl}

\end{document}

%% file: preamble.tex
\usepackage[T1]{fontenc}
\usepackage{booktabs}

\usepackage{microtype}

\makeatletter
\renewcommand\paragraph{\@startsection{paragraph}{4}{\z@}{.5em \@plus .2ex \@minus .1ex}{-1em}{\normalfont\normalsize\bfseries}}
\makeatother

\usepackage{xspace}
\newcommand{\method}{M2R\xspace}
\newcommand{\ourworlds}{MutableWorlds\xspace}

\usepackage{placeins}

\usepackage{colortbl}
\definecolor{rankfirst}{HTML}{FEB4B3}
\definecolor{ranksecond}{HTML}{FFD9B2}
\definecolor{rankthird}{HTML}{FEFFB3}
\newcommand{\first}[1]{\cellcolor{rankfirst}#1}
\newcommand{\second}[1]{\cellcolor{ranksecond}#1}
\newcommand{\third}[1]{\cellcolor{rankthird}#1}

\usepackage{pifont}

\usepackage{algorithm}
\usepackage{algpseudocode}

\usepackage{etoc}
\AtBeginDocument{\etocdepthtag.toc{main}}%
\newcommand{\cmark}{\textcolor{ForestGreen}{\ding{51}}}
\newcommand{\xmark}{\textcolor{BrickRed}{\ding{55}}}

%% file: sec/0_teaser.tex
\begin{center}
  \centering
  \captionsetup{type=figure}
  \includegraphics[width=\linewidth]{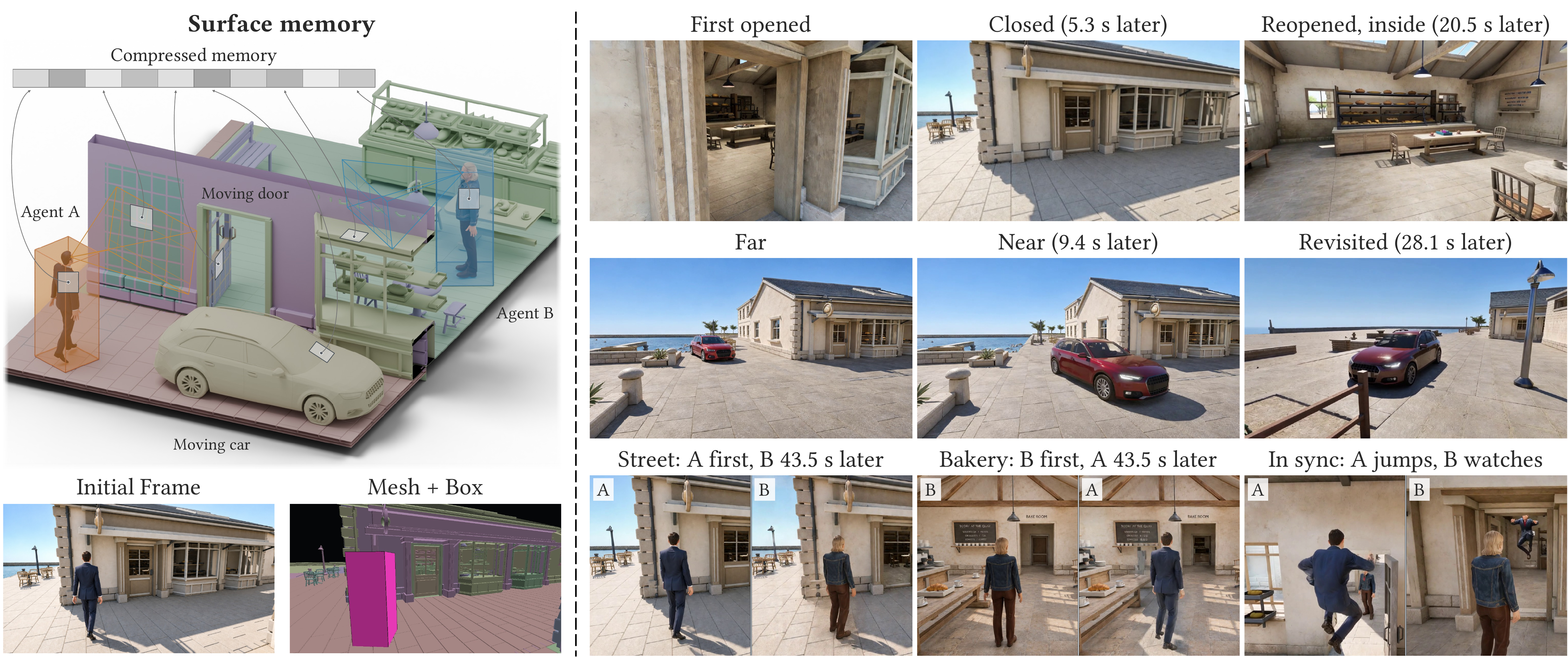}
  \captionof{figure}{Given an initial frame and a coarse, untextured mesh from a coding agent or a 3D scan, 
  \method renders photorealistic video for every agent in real time.
  It stores the features it generates on the mesh surfaces they depict, in a memory that all agents share (\textbf{left}).
  Appearance thus persists behind a reopened door, on a moving car, and across agents, long after leaving the model's context (\textbf{right}).
  }
  \label{fig:teaser}
\end{center}

%% file: sec/0_abstract.tex
\begin{abstract}
Video world models synthesize photorealistic observations in response to user actions but struggle to remain consistent over long interactions.
Coding agents and text-to-3D models instead generate explicit 3D worlds whose geometry and state persist by construction, but whose appearance falls far short of photorealism.
Recent methods combine the strengths of both by rendering such worlds into coarse proxies that guide a video model.
However, since these proxies only specify where things are, the renderer must invent what they look like as the world is explored, and anything it forgets is reinvented when a surface reappears.
Our key insight is that a mesh identifies which surface each pixel shows, across frames and views, even as objects move, allowing it to serve as an address for what the renderer generates.
We propose \method (Mesh to Remember), a few-step causal video renderer that stores its own features on the surfaces they depict, in each object's frame, and retrieves them by rasterizing the mesh whenever those surfaces come back into view.
Appearance thus travels with objects wherever they move, persists across revisits, and is shared by every agent that observes it, while memory grows with the surfaces seen instead of the number of frames.
Trained only on synthetic data, \method transfers zero-shot to real indoor scans and to worlds authored by coding agents.
It outperforms state-of-the-art methods in consistency across static, dynamic, and multi-agent benchmarks, more than doubling object permanence and nearly doubling cross-agent consistency, while rendering 720p video in real time (40~FPS).%
\end{abstract}

%% file: sec/1_intro.tex
\section{Introduction}
\label{sec:intro}

Interactive virtual worlds are essential for applications in gaming, embodied AI, and closed-loop simulation.
These use cases require both photorealistic rendering and persistent world state: a door opened by one agent should remain open when another walks by, and the room behind it should look the same each time it is opened (\cref{fig:teaser}).
In recent years, two paradigms have emerged that promise to build such worlds automatically: video world models and 3D world generators.

Video world models synthesize observations in response to user actions, from game worlds~\cite{bruce2024genie,valevski2025gamengen,he2025matrixgame2} to photorealistic scenes~\cite{he2025matrixgame2,nvidia2025cosmos,parkerholder2025genie3}, and increasingly for multiple players~\cite{enigma2025multiverse,savva2026solaris,wu2026multiworld,hu2026mira,liu2026gammaworld,huang2026multics}.
Despite this progress, these models must infer both how the world evolves and how it looks from their own past observations, and errors in either can accumulate over long interactions.
Memory mechanisms that retrieve past frames or their cached features, or maintain 3D caches, extend the history a model can draw upon and improve consistency when regions are revisited~\cite{xiao2025worldmem,sun2026worldplay,yi2026worldkv,wu2025spatialmemory,yu2026mosaicmem,shen2026lyra2}.
However, they tie memory to where content was observed, which no longer holds once the world changes.
When an object is carried across a room, its remembered appearance stays where the object used to be, and the model must reinvent it.

3D world generators instead create the world itself, from explorable scenes that text-to-3D models produce given a single prompt~\cite{hunyuanworld2025,hyworld2026} to entire dynamic worlds that coding agents write as programs, with articulated furniture~\cite{wang2026scenecode} and rules for how objects move~\cite{chen2026codeworld}.
As these worlds are explicit, their geometry and state persist by construction, and every agent observes the same world.
However, they fall short of photorealism,
as even the most sophisticated real-time game renderings remain distinguishable from real imagery~\cite{richter2023enhancing}.

Instead of treating these paradigms as standalone alternatives, recent methods combine their complementary strengths: the explicit world maintains persistent geometry and state, while a video model renders coarse proxies of it (geometry or semantic layouts) into realistic observations~\cite{huang2026programmable,chen2026codeworld,chi2026agentgarten,streamrenderh3_2026,liang2026codevideomodel,fang2026coder}.
Since these proxies only specify what is where, the renderer must invent the world's appearance as it is explored, and keep it consistent as agents revisit regions, move objects, and encounter surfaces that others have already seen.
A natural place to keep this appearance is the explicit world itself.
Existing methods paint generated colors back onto its geometry~\cite{richardson2023texture,hollein2023text2room,mallya2020worldconsistent,fang2026coder}, but colors are a lossy copy of what the renderer generated and must be re-rendered and re-encoded at every step, which is slow and accumulates errors (\cref{tab:abl-structure}).

We instead treat the mesh as an address.
Our key insight is that a mesh identifies the surface behind every pixel in every view, even after that surface has moved.
We therefore store the renderer's own features on the surfaces they depict as they are generated, and attend to them whenever those surfaces reappear.
This memory has three benefits.
First, as memory is indexed by surface instead of time, storage follows scene coverage and observations remain addressable long after leaving the model's context.
Second, when an object moves or articulates, its remembered appearance travels with it, whereas memory tied to camera poses or world positions stays behind.
Third, all agents read from and write to the same surfaces, allowing each to render what only others have seen.

We realize this insight in \method (Mesh to Remember), a real-time video world model that renders worlds of coarse, untextured meshes into photorealistic video from a single initial frame.
As users act and the world updates, \method conditions at every step on renderings of the current meshes and agent boxes from each agent's viewpoint, reusing stored appearance on observed surfaces and synthesizing newly revealed regions and view-dependent effects.
Agents are remembered the same way: \method generates each agent's body and motion from its bounding box, and the box carries that appearance with it.
We fine-tune a pretrained video diffusion model only on synthetic multi-agent episodes rendered in Unreal Engine with Isekai3D~\cite{karp2026isekai3d} and adapt it into a few-step causal renderer with rollout-based training~\cite{huang2025selfforcing}.
As this renderer remembers nothing outside its short context, we train it to write its own features to the surface memory during rollouts and to read them back on palindromic paths that revisit earlier views, so that it learns to use memory that carries its own errors.
In summary, our contributions are:
\begin{enumerate}
    \item We design a causal video renderer that conditions on renderings of scene geometry and agent bounding boxes, and reads and updates surface memory indexed by the mesh in each object's frame, so that remembered appearance follows objects as they move.
    \item We bound this memory by the observed surface area, compress it 18-fold with transform coding inspired by language-model KV caches~\cite{staniszewski2026kvtc}, and pair it with a fast video autoencoder, allowing \method to render at 40~FPS and keep 8{,}000-frame rollouts within 45~GiB of GPU memory.
    \item Although trained only on synthetic data, \method transfers zero-shot to real indoor scans and to articulated worlds authored by coding agents, and outperforms state-of-the-art methods on revisit, interaction, and multi-agent benchmarks.
\end{enumerate}

%% file: sec/2_related.tex
\begin{figure*}[t]
    \centering
    \includegraphics[width=\linewidth]{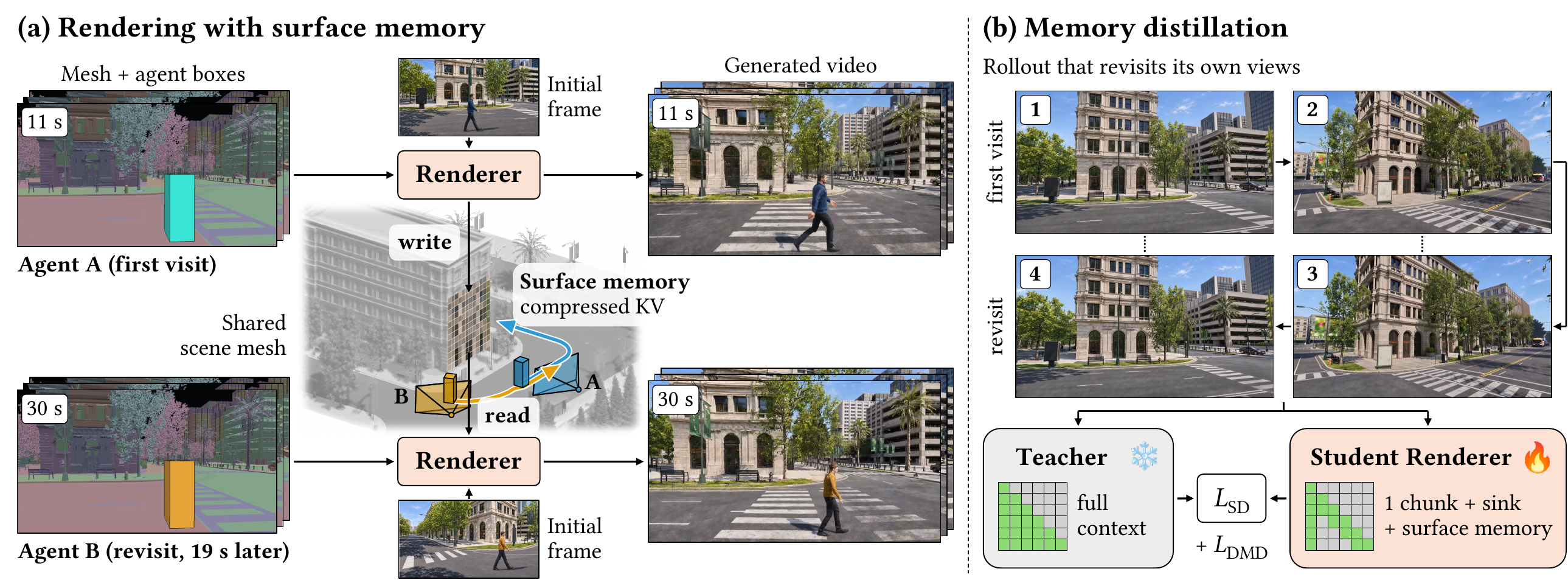}
    \caption{\textbf{Method overview.} \method renders each agent's view from the initial frame and renderings of a shared scene mesh.
    As agent~A explores, it stores generated keys and values in a surface memory, indexed by the mesh surfaces they depict.
    When agent~B encounters the same surfaces, it retrieves and renders them as agent~A saw them (\textbf{a}).
    We train our renderer on rollouts that revisit earlier views.
    With a one-chunk window, it must render each revisit from the memory it wrote, supervised by a frozen teacher that sees the full past (\textbf{b}).
  }
\label{fig:method}
\vspace{-8pt}
\end{figure*}
    
\section{Related Work}
\label{sec:related}

\paragraph{Interactive video world models.}
Interactive video world models~\cite{bruce2024genie,valevski2025gamengen,he2025matrixgame2,parkerholder2025genie3} are progressing from a single player to several~\cite{enigma2025multiverse,wu2026multiworld}, who must all observe the same evolving world.
Most multi-agent models keep this shared world inside the network, generating all players' views jointly~\cite{savva2026solaris,hu2026mira,liu2026gammaworld,mo2026worldweaver,huang2026multics}.
Joint generation grows costly as players are added~\cite{liu2026gammaworld,huang2026multics}, and the shared world can drift once players leave each other's view~\cite{savva2026solaris}.
Other methods render each view from an explicit state, such as an editable level map~\cite{po2026multigen}, an authoritative game state~\cite{cai2026mass}, an entity table~\cite{zhao2026khora}, or a shared point cloud~\cite{hu2026metaworld}.
While this keeps geometry and state consistent across agents, it does not retain the appearance that each view generates.
To share this appearance, recent methods exchange generated frames between agents~\cite{zhu2026shareverse,ye2026worldcast,xu2026consistworld}.
\method instead stores it directly on the mesh surfaces that every agent observes.

\paragraph{Generative rendering of explicit geometry.}
Generative models increasingly render explicit geometry into realistic footage, from G-buffers, depth, and segmentation maps~\cite{liang2025diffusionrenderer,nvidia2025cosmostransfer1} to renderings of animated meshes, low-poly games, and semantic block worlds~\cite{chen2026dar,cao2026fourdirector,gomeznogales2026coarsetoreal,hao2021gancraft}, some in real time~\cite{zhan2026magpie,nvidia2026dlss5}.
Executable worlds apply them to scenes written as programs, often by coding agents, with their geometry, state, and rules~\cite{chen2026codeworld,liang2026codevideomodel,chi2026agentgarten,huang2026programmable,streamrenderh3_2026,zhang2026matrixgame4}.
However, as geometry alone does not determine materials, colors, or identities~\cite{chi2026agentgarten}, the renderer must remember the appearance it invents.
Offline methods store this appearance in the 3D representation itself, in mesh textures~\cite{richardson2023texture,hollein2023text2room}, in the UV space of animated meshes~\cite{cai2024generativerendering}, or by distilling refined views into captured scenes~\cite{wu2025difix3d,delutio2026artifixer}.
They require the animation or target views in advance, which an interactive world cannot provide.
Closest to our renderer, the concurrent AgentGarten~\cite{chi2026agentgarten} adapts the same backbone into a few-step causal renderer of code-defined worlds.
While it keeps appearance only in a reference frame and its recent context, and can lose it after long occlusions, \method stores it on the mesh itself and thus remains consistent over long rollouts (\cref{sec:exp-dynamic}).
\looseness=-1 %

\paragraph{Memory for long video generation.}
Most memories retrieve past frames by camera pose or view overlap~\cite{xiao2025worldmem,yu2025contextasmemory,sun2026worldplay,wang2026matrixgame3} or return the cached keys and values of past chunks~\cite{hong2025relic,yi2026worldkv,chen2026reworld,zhang2026mosaichunk}.
Unless bounded~\cite{chen2026reworld,dou2026slotmemory}, these memories grow with the rollout, reaching an estimated 300~GiB of CPU memory per thousand frames for the concurrent MosaiChunk's image-to-video backbone~\cite{zhang2026mosaichunk}.
They also leave the network to infer how the retrieved content corresponds to the current view.
Geometric memories make this correspondence explicit by storing observations in world-fixed point clouds, surfels, or per-frame geometry~\cite{mallya2020worldconsistent,ren2025gen3c,huang2025memoryforcing,li2025vmem,wu2025spatialmemory,shen2026lyra2}, lifting features or latent patches into world space~\cite{zhou2025persistent,yu2026mosaicmem,qian2026matrixgame35,wang2026lsmworld}, or reprojecting tokens with engine depth~\cite{ma2026closingloop}.
However, this correspondence only holds for static content, as moving objects are filtered out~\cite{wu2025spatialmemory,wang2026lsmworld,qian2026matrixgame35} or left to the generator~\cite{yu2026mosaicmem}, and even TourPhysics~\cite{zhang2026tourphysics}, which routes the generator's own attention values with simulator geometry, excludes them from long-term memory.
Memories that do handle motion replay it from an input video~\cite{ren2025gen3c,zhu2026demiurge}, predict it while objects are out of sight~\cite{chen2026hydra,duan2026liveworld,shin2026videoworldmodelstrack}, or remember objects by identity~\cite{xiong2026actworld,dou2026slotmemory,wang2026worlddirector}, but none track which of an object's surfaces is visible.
Closest to our memory, the concurrent CoDeR~\cite{fang2026coder} registers generated colors on whitebox surfaces, where later views and other agents observe them.
As it renders 124 frames at a time, however, the effect of an action becomes visible to any agent only after a full clip is rendered.
It also learns to read colors registered from its training targets instead of its own outputs, which lets its errors accumulate over long rollouts.
\method instead keeps the renderer's own keys and values, which it writes after every chunk and learns to read back from its own rollouts.
Stored in mesh surface cells in each object's frame, they follow objects as they move, grow only with the surfaces seen, and can be read by any agent in real time.

%% file: sec/3_method.tex
\section{Preliminaries}
\label{sec:preliminaries}
\paragraph{Video diffusion models.}
We build upon video diffusion models, which typically operate in the latent space of a video autoencoder.
Flow matching~\cite{lipman2023flowmatching} learns to transport Gaussian noise into these latents.
Given clean latents $\mathbf{z}$, noise $\boldsymbol{\epsilon}\sim\mathcal{N}(0,I)$, and flow time $\tau\in[0,1]$, we construct $\mathbf{z}_{\tau}=(1-\tau)\boldsymbol{\epsilon}+\tau\mathbf{z}$ and train a velocity field $v_\theta$ by minimizing $\mathbb{E}_{\mathbf{z},\boldsymbol{\epsilon},\tau}\|v_\theta(\mathbf{z}_{\tau},\tau;\mathbf{c})-(\mathbf{z}-\boldsymbol{\epsilon})\|_2^2$, where $\mathbf{c}$ denotes conditioning signals such as actions or renderings of scene geometry.
At inference, integrating the field from $\tau=0$ to $\tau=1$ produces clean latents.

\paragraph{Autoregressive video generation.}
As bidirectional video models denoise all frames of a clip jointly, their memory cost grows with the number of frames.
To solve this, long-horizon interactive world models instead generate video autoregressively, one latent chunk at a time.
Each chunk attends to the chunks before it~\cite{yin2025causvid}, whose keys and values are kept in a KV cache.
To keep this cache bounded, only a sliding window of recent chunks is typically retained~\cite{huang2025selfforcing,shin2026motionstream}, at the cost of reinventing any content that reappears after leaving the window.
As autoregressive models condition on their own outputs at inference, Self Forcing~\cite{huang2025selfforcing} trains them on their own rollouts, supervised by distribution matching distillation (DMD)~\cite{yin2024dmd} from a bidirectional teacher.

\begin{figure}[t]
  \centering
  \includegraphics[width=\linewidth]{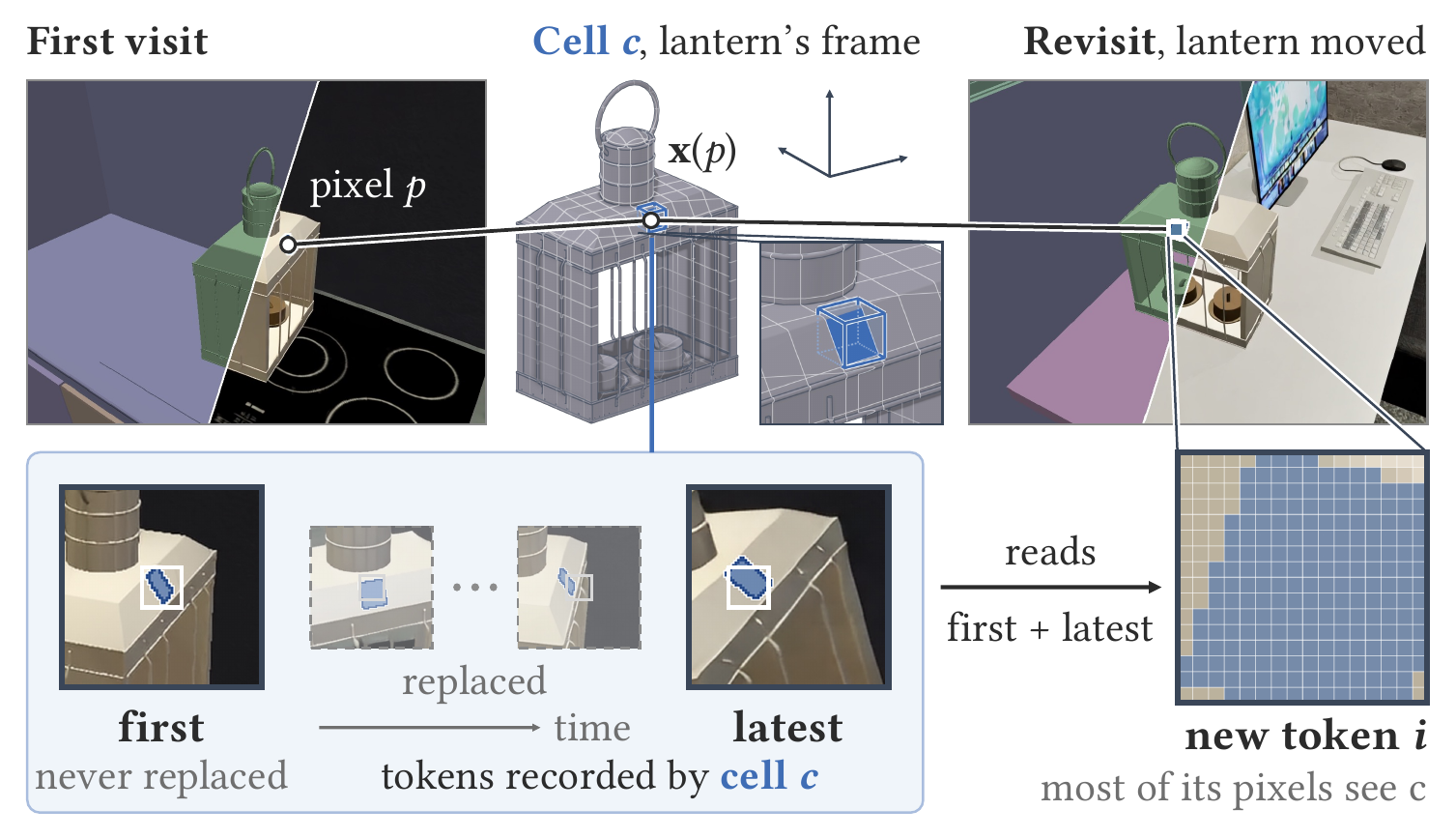}
  \caption{\textbf{Surface memory.}
  We assign each pixel $p$ to the cell $c$ that contains its surface point $\mathbf{x}(p)$ in the object's coordinate frame (\textbf{top}).
  After each chunk, we record the token that best covers each visible cell, keeping only the first and latest (\textbf{bottom left}).
  When we revisit the lantern after it has moved, each new token $i$ reads the past tokens that most of its pixels point to (\textbf{bottom right}).}
  \label{fig:surface_memory}
\vspace{-4pt}
\end{figure}

\section{Method}
\label{sec:method}
Given a coarse, untextured mesh, a caption of the scene, and an initial first- or third-person frame for each agent, our goal is to render consistent, photorealistic views in real time as agents explore and change the world.
We do so via a few-step causal video renderer that is conditioned on renderings of the mesh and stores generated features in a surface memory shared by all agents (\cref{fig:method}).
As the mesh identifies the surface behind every pixel, the renderer can retrieve these features whenever the surface they depict reappears, even after it has moved.

\subsection{Geometry Conditioning}
\label{sec:method-geometry}
\label{sec:method-rendering}
To specify what is visible in each chunk, we rasterize the current mesh and agent boxes from each agent's camera into a proxy video, with flat shading and one arbitrary color per asset to separate objects.
We encode it into the backbone's latent space and add it to the embedded noisy latents through a zero-initialized patch embedding, which keeps the sequence length unchanged.

\subsection{Surface Memory}
\label{sec:method-memory}

Most autoregressive video models remember between 24 and 72 frames in a key-value cache~\cite{huang2025selfforcing,yang2025longlive,shin2026motionstream}.
Instead, we pair a much smaller temporal cache (8 frames) with a surface memory shared across agents, which indexes keys and values by the mesh surfaces they depict.
Once observed surfaces are evicted from the temporal cache, they remain retrievable by any agent that later sees them, and the memory grows with the surfaces observed instead of with time (\cref{fig:persistence}b).

\begin{figure}[t]
  \centering
  \includegraphics[width=\linewidth]{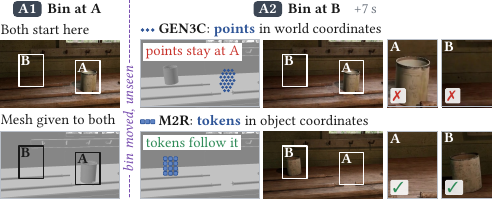}
  \caption{\textbf{Moving objects.} A bin is seen at A (A1) and then moved to B while the camera looks away.
  GEN3C~\cite{ren2025gen3c} indexes its 3D cache by world position, which leaves the bin's appearance at A and nothing at B (\textbf{top}), and thus still renders the bin at A and none at B when the camera looks back (A2).
  As we define cells in each object's frame, the bin keeps its cells (\textbf{bottom}), and \method renders it at B and no longer at A.}
  \label{fig:moving}
\vspace{-8pt}
\end{figure}

\paragraph{Cells.} To index the memory, we divide the surfaces of the mesh and agent boxes into small cells.
We define cells in the coordinates of the object they lie on, which lets a cell follow its surface as the object moves or articulates, with no update to the memory (\cref{fig:moving}).
Rasterizing this geometry into a view gives the object $o(p)$ and the surface point $\mathbf{x}(p)$ behind each pixel $p$, and we assign the pixel to the cell that holds this point (\cref{fig:surface_memory}, top),
\begin{equation}
  c(p) = \Big(o(p),\; \big\lfloor \mathbf{T}_{o(p)}(t)\,\mathbf{x}(p) / s \big\rfloor\Big),
  \label{eq:cell}
\end{equation}
where $\mathbf{T}_{o}(t)$ maps world coordinates into the coordinates of object $o$ at time $t$ and $s$ is the cell size.
As cells do not depend on any camera, all agents address a surface through the same cell.
In practice, we also split cells by the direction the surface faces and keep them at four sizes, so that surfaces seen from afar still fill the coarser cells (\cref{sec:supp-memory}).

\begin{table*}[t]
  \centering
  \caption{\textbf{Static environments.}
  We report Orbia scores~\cite{lu2026orbia} (0--100, higher is better) on ScanNet++ and Replica, with the {\fboxsep=1pt \colorbox{rankfirst}{best}, \colorbox{ranksecond}{second}, and \colorbox{rankthird}{third}} results highlighted.
  Throughout the tables, $\dagger$ and $\S$ mark methods that generate at 480p and 540p, while all others generate at 720p.
  \method outperforms all baselines on every metric except visual quality, and is the only method to render 720p video in real time.
  }
  \label{tab:static}
  \footnotesize
  \setlength{\tabcolsep}{1.6514pt}%
  \newcommand{\hcell}[1]{\begin{tabular}[c]{@{}c@{}}#1\end{tabular}}
  \resizebox{\linewidth}{!}{%
  \begin{tabular}{@{}lccccccccccccccccc}
    \toprule
    & \multicolumn{6}{c}{ScanNet++~\cite{yeshwanth2023scannetpp}} & \multicolumn{6}{c}{Replica~\cite{straub2019replica}} & \multicolumn{5}{c}{Efficiency} \\
    \cmidrule(lr){2-7}\cmidrule(lr){8-13}\cmidrule(lr){14-18}
    Method & \hcell{Control\\Align.} & \hcell{Visual\\Quality} & \hcell{Input\\Pres.} & \hcell{Content\\Persist.} & \hcell{3D\\Consist.} & \hcell{Temp.\\Stab.} & \hcell{Control\\Align.} & \hcell{Visual\\Quality} & \hcell{Input\\Pres.} & \hcell{Content\\Persist.} & \hcell{3D\\Consist.} & \hcell{Temp.\\Stab.} & \hcell{\#\\Params.} & \hcell{Lat.\\(s)$\downarrow$} & FPS$\uparrow$ & MP/s$\uparrow$ & \hcell{Peak\\(GiB)$\downarrow$} \\
    \midrule
    SolarWM H3~\cite{huang2026solarwm} & 83.01 & 71.72 & 71.46 & 80.34 & 74.96 & 87.22 & 84.26 & \first{73.65} & 74.23 & 84.09 & 76.46 & 88.99 & 33B & 9.8 & 1.73 & 1.79 & 113.6 \\
    \midrule
    Matrix-Game~3.5~\cite{qian2026matrixgame35} & 62.72 & 69.52 & 71.18 & 54.58 & 54.15 & 75.89 & 68.67 & 72.21 & 69.77 & 66.23 & 56.25 & 79.07 & 5B & \third{4.5} & 2.67 & \third{2.41} & 63.6 \\
    GEN3C~\cite{ren2025gen3c} & 85.48 & 67.21 & 80.68 & \third{85.46} & 72.57 & 86.93 & \second{95.82} & 69.64 & \second{88.88} & \second{92.06} & 75.42 & 90.83 & 7B & 147.5 & 0.81 & 0.73 & \third{50.9} \\
    Lyra~2.0$^\dagger$~\cite{shen2026lyra2} & \second{92.21} & 71.71 & 82.26 & 81.92 & \third{75.18} & \third{89.88} & 93.33 & 72.05 & 83.57 & 86.26 & 76.21 & 90.43 & 14B & 170.2 & 0.47 & 0.19 & 72.6 \\
    Lyra~2.0 (DMD)$^\dagger$~\cite{shen2026lyra2} & \third{92.11} & 71.60 & \third{82.42} & 85.36 & \second{75.23} & \second{90.11} & \third{94.18} & 72.85 & 82.64 & 86.53 & 76.73 & \second{91.30} & 14B & 23.7 & \third{3.38} & 1.35 & 73.0 \\
    \midrule
    WorldDirector$^\dagger$~\cite{wang2026worlddirector} & 78.62 & 67.60 & 79.48 & 73.75 & 70.14 & 83.43 & 85.10 & 68.84 & 83.86 & 84.84 & 75.90 & 86.60 & 14B & 289.6 & 0.28 & 0.11 & 67.9 \\
    Code World Model~\cite{chen2026codeworld} & 83.64 & 71.93 & 82.40 & 80.14 & 74.41 & 89.03 & 85.89 & 71.14 & 83.82 & 76.81 & \third{77.23} & 88.68 & 33B & 535.7 & 0.17 & 0.18 & 75.7 \\
    Code Video Model$^\S$~\cite{liang2026codevideomodel} & 67.60 & \second{73.54} & 74.84 & 70.24 & 74.98 & 85.26 & 71.99 & 71.21 & 78.62 & 69.89 & 75.22 & 85.00 & 33B & 166.0 & 0.75 & 0.39 & 146.3 \\
    AgentGarten$^\dagger$~\cite{chi2026agentgarten} & 88.92 & \first{73.66} & \second{87.15} & \second{89.49} & 75.02 & 88.62 & 93.20 & \second{73.63} & \third{88.44} & \third{91.72} & \second{78.89} & \third{91.10} & 8B & \second{0.5} & \second{34.1} & \second{13.6} & \second{39.2} \\
    \midrule
    \method & \first{95.63} & \third{72.52} & \first{88.54} & \first{89.88} & \first{76.60} & \first{91.37} & \first{97.80} & \third{72.91} & \first{92.27} & \first{92.16} & \first{79.69} & \first{92.34} & 8B & \first{0.19} & \first{41.1} & \first{37.0} & \first{33.6} \\
    \bottomrule
  \end{tabular}}
\end{table*}

\paragraph{Writing.}
As the renderer operates on tokens that each span a patch of pixels across several frames, a single token often covers multiple cells.
We therefore store each token's keys and values once and let cells point to it.
After generating chunk $n$, we associate each visible cell $c$ with the token that covers most of its pixels, $j = \arg\max_{j'} |\{p \in \Omega_{j'} : c(p) = c\}|$, where $\Omega_{j'}$ denotes the pixels of token $j'$, and record the observation $(n, j)$ (\cref{fig:surface_memory}, bottom left).
We keep two observations per cell: the first, which we never replace~\cite{xu2026recapforcing} to prevent errors from accumulating across revisits, and the latest, which reflects the surface's current state.

\paragraph{Storage.}
Storing the keys and values of every token is expensive~\cite{xi2026quantvideogen,yi2026worldkv}, amounting to about 54~GiB per minute of 720p video at 30~FPS.
To limit this cost, we discard tokens as they leave the temporal cache unless a cell still points to them.
We further compress the remaining tokens following key-value compression for language models~\cite{staniszewski2026kvtc,chang2026xkv}.
We project each token's keys and values across all layers onto a principal component basis that we fit once on the renderer's own rollouts, and store the result at 8-bit precision.
This shrinks each token 18-fold without degrading consistency (\cref{sec:supp-memory,tab:supp-compression}).
The temporal cache, which also keeps the initial chunk as an attention sink~\cite{xiao2024attentionsinks,yang2025longlive,liu2025rollingforcing}, stays uncompressed.

\paragraph{Reading.}
For each token $i$ of a new chunk, we look up the first and latest observations of the cells that it covers.
Since a token can cover multiple cells, we select the past token that most of its pixels $p \in \Omega_i$ point to, choosing first its chunk $n_i$ and then the token $j_i$ within that chunk,
\begin{equation}
\begin{aligned}
  n_i &= \operatorname*{arg\,max}_{n}\, \big|\{\, p \in \Omega_i : (n, \cdot) \in \mathcal{O}(p) \,\}\big|,\\
  j_i &= \operatorname*{arg\,max}_{j}\, \big|\{\, p \in \Omega_i : (n_i, j) \in \mathcal{O}(p) \,\}\big|,
\end{aligned}
\label{eq:read-vote}
\end{equation}
where $\mathcal{O}(p)$ is the first observation of the cell at pixel $p$.
We repeat this selection with the latest observations and retrieve a set $\mathcal{R}_i$ of at most two past tokens (\cref{fig:surface_memory}, bottom right).

\paragraph{Memory conditioning.}
Each token attends to the tokens retrieved for its chunk in addition to the temporal cache.
We assign each token in $\mathcal{R}_i$ the spatial position of token $i$, as in the re-positioned frame and patch memories of recent world models~\cite{sun2026worldplay,yu2026mosaicmem,qian2026matrixgame35,xu2026ucm} (\cref{sec:supp-memory,fig:supp-read-time}).
As each token retrieves at most two past tokens, the cost of rendering stays independent of the length of the interaction.

\paragraph{Multiple agents.}
With several agents, each agent keeps its own temporal cache, while all agents read from and write to the same surface memory.
Each agent can thus render surfaces that only others have seen (\cref{tab:multiagent}), even though we train with a single view at a time (\cref{sec:supp-memory}).

\begin{figure*}[t]
  \centering
  \includegraphics[width=\linewidth]{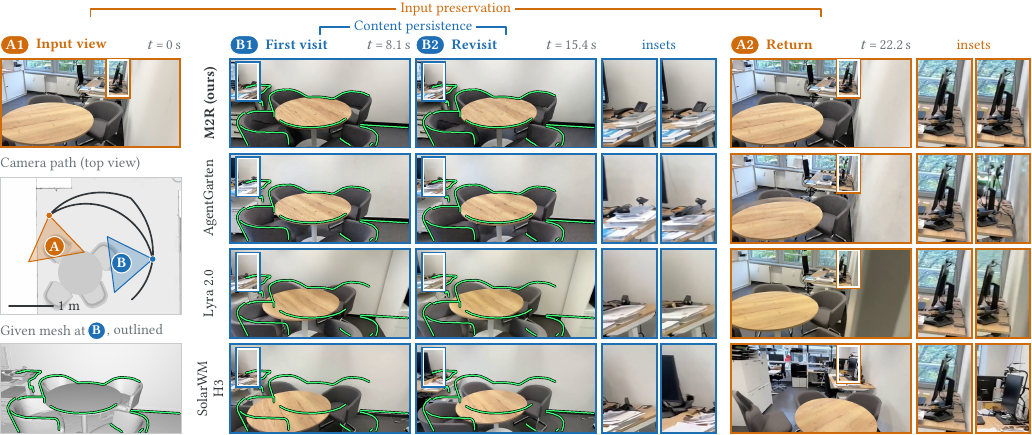}
  \caption{\textbf{Static environments.} We show a ScanNet++ rollout that returns to the input view (\textbf{A}) and revisits a generated view (\textbf{B}).
  Lyra~2.0 and SolarWM H3 fail to preserve the input view (\textbf{bottom rows}), and AgentGarten alters the monitor by the window (\textbf{right insets}).
  Although trained only on synthetic data, \method is the only method to preserve both views.
  }
  \label{fig:static}
  \vspace{-8pt}
\end{figure*}

\subsection{Optimization}
\label{sec:method-training}

\paragraph{Causal renderer.}
We start from a bidirectional image-to-video model and fine-tune it with flow matching on videos paired with mesh renderings and keyboard controls (\cref{sec:supp-rendering,sec:supp-optimization}), so that it learns to follow the geometry.
For real-time generation, we make it causal and few-step following Causal Forcing++~\cite{zhao2026causal}, and then train it on its own rollouts with Self Forcing (\cref{sec:preliminaries}) to limit how errors compound over long videos.
The resulting renderer generates arbitrarily long videos with a few sampling steps per chunk, but remembers nothing outside its temporal cache.

\paragraph{Memory distillation.}

Although the causal renderer already benefits from the memory without any training (\cref{sec:supp-ablations,tab:supp-compression}), it has little incentive to rely on it.
We therefore apply Self Forcing a second time, except that the renderer now reads from and writes to its memory on rollouts that revisit earlier views (\cref{fig:method}b).
We construct these rollouts as palindromes, in which the camera goes back and forth up to four times, such that each return can only be rendered from memory (as the temporal cache holds a single chunk).
To supervise each return, we use a frozen copy of the causal renderer that attends to the entire rollout (and thus still sees the first visit) as a teacher~\cite{liu2026opsdv,zhang2026mosaichunk}, and minimize the difference between their clean-latent predictions $\hat{\mathbf{z}}^{T}$ and $\hat{\mathbf{z}}^{S}$ at every chunk:
\begin{equation}
  \mathcal{L}_{\mathrm{SD}}
  = \mathbb{E}_{n,\tau}
  \left[\left\|\hat{\mathbf{z}}^{S}_{n,\tau}
  -\operatorname{sg}\!\left(\hat{\mathbf{z}}^{T}_{n,\tau}\right)
  \right\|_2^2\right],
  \label{eq:memory-distillation}
\end{equation}
where $\operatorname{sg}$ stops gradients.
Our final loss is $\mathcal{L} = \mathcal{L}_{\mathrm{DMD}} + \lambda\,\mathcal{L}_{\mathrm{SD}}$, with $\lambda = 0.1$ (\cref{sec:supp-optimization,fig:supp-sd-weight}).

%% file: sec/4_experiments.tex
\section{Experiments}
\label{sec:experiments}

We evaluate \method in the three situations in which an interactive world model must remember appearance: a camera revisits a region, an object moves, and several agents see the same surfaces.

\subsection{Implementation}
\label{sec:implementation}
We use Cosmos3-Nano~\cite{nvidia2026cosmos3} as the base model and train on 64 GPUs with paired mesh and RGB videos from Isekai3D~\cite{karp2026isekai3d}.
At inference time, we generate eight-frame chunks (two latent frames) with two-step denoising, and measure all timings with a single GB200.
To speed up encoding and decoding, we replace the base model's VAE with FastVAE, a transformer autoencoder that we train from scratch for the same latent space.

\input{sec/tab_dynamic_one_column}

\subsection{Static Environments}
\label{sec:exp-static}

\paragraph{Datasets.}
We evaluate on 46 scenes of ScanNet++~\cite{yeshwanth2023scannetpp}, real indoor captures with laser-scanned meshes, and on the 18 scenes of Replica~\cite{straub2019replica}, whose dense meshes provide ground truth from any viewpoint.
We follow the revisit protocol of Orbia~\cite{lu2026orbia}, in which each 22-second trajectory starts at an input view, discovers a region absent from it, explores elsewhere, and then revisits both.
We use Orbia's trajectories on ScanNet++ and generate four per scene on Replica (\cref{sec:supp-static}).
\paragraph{Baselines.}

We compare \method to three groups of baselines (\cref{tab:static}), namely a video world model that takes no geometry~\cite{huang2026solarwm}, methods that store what they generate in a 3D cache~\cite{qian2026matrixgame35,ren2025gen3c,shen2026lyra2}, and renderers that condition on proxy videos of the geometry~\cite{wang2026worlddirector,chen2026codeworld,liang2026codevideomodel,chi2026agentgarten}.
For fairness, we provide the ground-truth mesh to all methods that take geometry.

\begin{figure*}[t]
  \centering
  \includegraphics[page=1,width=\linewidth]{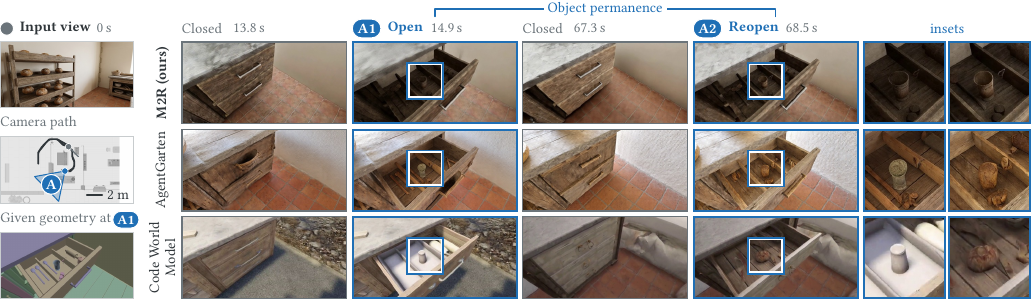}\\[5pt]
  \includegraphics[page=2,width=\linewidth]{figures/gallery_our_worlds_v1_compact.pdf}
  \caption{\textbf{Dynamic environments.} We show two rollouts in \ourworlds in which a drawer is reopened (\textbf{top}) and a lamp is moved while out of view (\textbf{bottom}).
  AgentGarten and Code World Model replace the cup in the drawer with a loaf (\textbf{top insets}) and fail to preserve the lamp across visits.
  \method is the only method to preserve appearance when drawers reopen and when objects move.
  }
  \label{fig:dynamic}
\end{figure*}

\begin{figure}[t]
  \centering
  \includegraphics[width=0.95\linewidth]{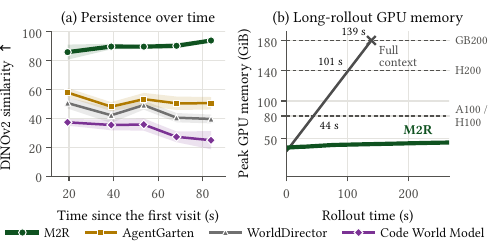}
  \caption{\textbf{Persistence and memory over time.}
  We plot DINOv2 similarity between the first and later openings of drawers and doors on HSSD (\textbf{a}) and peak GPU memory over an 8{,}001-frame rollout (\textbf{b}).
  Unlike the baselines, our memory does not fade with time.
  \method stays below 45~GiB, whereas attending to every past chunk runs out of memory after 139~s.}
  \label{fig:abl-plots}
  \label{fig:persistence}
  \vspace{-12pt}
\end{figure}

\begin{table}[t]
  \centering
  \caption{\textbf{Multi-agent environments.}
  We render two or four agents that share a world in \ourworlds, each from its own third-person camera.
  \method is the only method to share appearance between agents, and outperforms all baselines by a wide margin.}
  \label{tab:multiagent}
  \footnotesize
  \setlength{\tabcolsep}{3.0489pt}%
  \newcommand{\hcell}[1]{\begin{tabular}[c]{@{}c@{}}#1\end{tabular}}
  \begin{tabular}{@{}lcccc}
    \toprule
    & \multicolumn{2}{c}{Two agents} & \multicolumn{2}{c}{Four agents} \\
    \cmidrule(lr){2-3}\cmidrule(lr){4-5}
    Method & \hcell{Cross-Agent\\Consist.} & \hcell{Agent\\Persist.} & \hcell{Cross-Agent\\Consist.} & \hcell{Agent\\Persist.} \\
    \midrule
    SolarWM H3~\cite{huang2026solarwm} & 27.10 & 1.52 & 25.38 & 0.36 \\
    \midrule
    Matrix-Game~3.5~\cite{qian2026matrixgame35} & 27.65 & 0.00 & 31.81 & 0.11 \\
    GEN3C~\cite{ren2025gen3c} & 26.82 & 4.68 & 24.80 & 1.80 \\
    Lyra~2.0$^\dagger$~\cite{shen2026lyra2} & 21.70 & 15.19 & 22.20 & 13.96 \\
    Lyra~2.0 (DMD)$^\dagger$~\cite{shen2026lyra2} & 26.05 & 4.95 & 23.33 & 2.39 \\
    \midrule
    WorldDirector$^\dagger$~\cite{wang2026worlddirector} & 26.07 & \second{52.76} & 27.29 & \second{36.98} \\
    Code World Model~\cite{chen2026codeworld} & \third{36.51} & \third{43.08} & \third{37.45} & \third{31.41} \\
    Code Video Model$^\S$~\cite{liang2026codevideomodel} & 32.05 & 12.62 & 25.38 & 22.94 \\
    AgentGarten$^\dagger$~\cite{chi2026agentgarten} & \second{42.60} & 22.43 & \second{42.16} & 29.74 \\
    \midrule
    \method & \first{79.99} & \first{87.25} & \first{77.12} & \first{73.69} \\
    \bottomrule
  \end{tabular}
  \vspace{-8pt}
\end{table}

\begin{figure*}[t]
  \centering
  \includegraphics[width=0.98\linewidth]{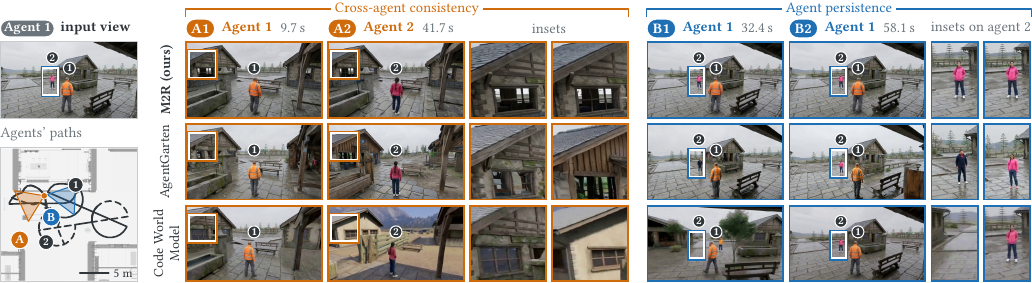}
  \caption{\textbf{Multiple agents.} We show a rollout in \ourworlds in which two agents look from the same viewpoint (\textbf{A}) and meet twice (\textbf{B}).
  AgentGarten and Code World Model show agent 2 a different building (\textbf{A2}) and fail to preserve agent 2 across meetings (\textbf{B}).
  \method is the only method to share appearance between agents.}
  \label{fig:multiagent}
  \vspace{-8pt}
\end{figure*}

\paragraph{Metrics.}
We report the six scores of Orbia~\cite{lu2026orbia} using its released evaluator: control alignment, visual quality, 3D consistency, temporal stability, and two memory scores.
\emph{Input preservation} compares later views of the surfaces visible in the input with ground truth, and \emph{content persistence} compares each revisit of the discovered region with its first generated view.
We also measure latency (from an input to the first frame it affects), frame rate, throughput in megapixels per second (MP/s), and peak GPU memory.

\paragraph{Results.}

We present quantitative results in \cref{tab:static} and qualitative examples in \cref{fig:static}.
Lyra~2.0 and SolarWM H3 lose the input view, and AgentGarten alters revisited objects.
Although trained only on synthetic data, \method outperforms all baselines on every metric except visual quality, where it performs competitively with the concurrent AgentGarten~\cite{chi2026agentgarten} and Code Video Model~\cite{liang2026codevideomodel}.
It is also the only method to render 720p video in real time, with 2.7$\times$ the throughput of AgentGarten and over 15$\times$ that of every other baseline.

\begin{table}[t]
  \centering
  \caption{\textbf{Diagnostics.} 
  We evaluate content persistence on ScanNet++, revisit on HSSD, and peak GPU memory over a 357-frame rollout.
  All trained variants, including \method, use 16 GPUs for 300 iterations.
  Storing the renderer's key-value (K/V) features instead of colors is crucial to rendering revisits consistently, and compression shrinks the additional K/V store from 3.70 to 0.21~GiB ($18\times$).
  }
  \label{tab:diagnostics}
  \label{tab:abl-content}
  \label{tab:abl-structure}
  \label{tab:abl-training}
  \footnotesize
  \setlength{\tabcolsep}{1.6pt}
  \newcommand{\hcell}[1]{\begin{tabular}[c]{@{}c@{}}#1\end{tabular}}
  \begin{tabular}{@{}lcccc@{\hskip 4pt}cccc@{}}
    \toprule
    Memory & Compr. & \hcell{Revisiting\\Rollouts} & \hcell{Self-\\Distill.} & \hcell{Fast\\VAE} & \hcell{Content\\Persist.} & Revisit & \hcell{Peak\\(GiB)$\downarrow$} & FPS$\uparrow$ \\
    \midrule
    None & -- & -- & -- & \cmark & 84.45 & 74.33 & 31.7 & 51.1 \\
    \midrule
    RGB & -- & \cmark & \cmark & \cmark & 83.59 & 79.56 & 32.9 & 19.8 \\
    \midrule
    K/V & \xmark & \cmark & \cmark & \cmark & 89.74 & 85.73 & 38.2 & 42.9 \\
    K/V & \cmark & \xmark & \cmark & \cmark & 89.09 & 84.47 & 33.6 & 41.1 \\
    K/V & \cmark & \cmark & \xmark & \cmark & 89.53 & 83.73 & 33.6 & 41.1 \\
    K/V & \cmark & \cmark & \cmark & \xmark & 89.76 & 86.52 & 42.7 & 19.2 \\
    \midrule
    \method & \cmark & \cmark & \cmark & \cmark & 89.83 & 86.78 & 33.6 & 41.1 \\
    \bottomrule
  \end{tabular}
\vspace{-4pt}
\end{table}

\subsection{Dynamic and Multi-Agent Environments}
\label{sec:exp-dynamic}

\paragraph{Datasets and baselines.}

We evaluate on 45 trajectories in five articulated HSSD homes~\cite{khanna2024hssd,chang2025partnr} and on 52 in \ourworlds, eight worlds that we build with a coding agent.
The camera returns to rooms it has seen, reopens drawers and doors with objects inside, and revisits objects that have moved.
We compare to the same methods as in \cref{sec:exp-static}.

\paragraph{Metrics.}
We report control alignment and visual quality as in \cref{sec:exp-static}.
We measure memory by comparing a method's frames at each revisit with its own frames from the same camera poses on the first visit.
\emph{Revisit} compares the static parts of a view, and \emph{object permanence} compares the \mbox{contents} of each reopened drawer or door.
\emph{Moved objects} checks that an object appears only at its current location, and that its appearance has not changed.
We compute these scores from DINOv2 features~\cite{oquab2024dinov2} and ratings by Gemini~3.7 Flash~\cite{google2026gemini37flash}, as detailed in \cref{sec:supp-dynamic}.

\paragraph{Results.}
We present quantitative results in \cref{tab:dynamic} and qualitative examples in \cref{fig:dynamic}.
\method outperforms all baselines on every metric except visual quality, where it is second only to the concurrent AgentGarten.
However, AgentGarten keeps appearance only in a reference frame and its recent context, and often shows new contents when a drawer or door is reopened (\cref{fig:dynamic}).
We more than double its object permanence (91.5 vs.\ 34.7 on HSSD and 94.1 vs.\ 43.4 in \ourworlds).
Our memory does not fade with the time between visits (\cref{fig:persistence}a).
GEN3C and Lyra~2.0 index their 3D caches by world position, which leaves a moved object's appearance where it was first seen (\cref{fig:moving}).
\method is the only method to preserve appearance when drawers reopen and when objects move.

\paragraph{Multiple agents.}
In \ourworlds, we also render 30 two-agent and 23 four-agent episodes, in which each agent later sees what another agent saw first.
\emph{Cross-agent consistency} averages revisit, object permanence, and moved objects when the first visit comes from another agent's video, and \emph{agent persistence} compares another agent's appearance across two separate meetings.
The baselines generate each agent's video separately, and thus give a new appearance to what another agent has already seen.
We nearly double the best cross-agent consistency among them (80.0 vs.\ 42.6 with two agents and 77.1 vs.\ 42.2 with four) and raise the best agent persistence from 52.8 to 87.3 with two agents and from 37.0 to 73.7 with four (\cref{tab:multiagent,fig:multiagent}).
\method is the only method to share appearance between agents.

\subsection{Diagnostics}
\label{sec:diagnostics}

We ablate what the memory stores, its compression, revisiting rollouts, the self-distillation loss, and FastVAE (\cref{tab:diagnostics}).
Storing the renderer's keys and values instead of colors is crucial, as colors on the same surface cells are less consistent (79.6 vs.\ 86.8 revisit) and must be re-rendered and re-encoded for every chunk, which halves the frame rate.
Retrieving latent frames by view frustum instead renders at 15~FPS, and is less consistent on revisits and moved objects in dynamic scenes (\cref{sec:supp-ablations,tab:supp-structure,fig:supp-frustum}).
Compression shrinks the additional K/V store 18-fold at no loss and allows rollouts of over 8{,}000 frames within 45~GiB (\cref{fig:persistence}b).  
Revisiting rollouts and the self-distillation loss each improve revisit, and FastVAE doubles the frame rate.

%% file: sec/tab_dynamic_one_column.tex
\begin{table}[t]
  \centering
  \caption{\textbf{Dynamic environments.}
  We report control alignment (Ctrl.), visual quality (Qual.), object permanence (Perm.), revisit (Rev.), and moved objects (0--100, higher is better).
  \method more than doubles the best baseline's object permanence, and outperforms all baselines on every metric except visual quality.}
  \label{tab:dynamic}
  \scriptsize
  \setlength{\tabcolsep}{1.2pt}%
  \begin{tabular}{@{}lcccccccccc}
    \toprule
    & \multicolumn{5}{c}{HSSD~\cite{khanna2024hssd}} & \multicolumn{5}{c}{\ourworlds} \\
    \cmidrule(lr){2-6}\cmidrule(lr){7-11}
    Method & Ctrl. & Qual. & Perm. & Rev. & Moved & Ctrl. & Qual. & Perm. & Rev. & Moved \\
    \midrule
    SolarWM H3 & \third{50.6} & 61.2 & 14.0 & 33.1 & 31.7 & \third{70.3} & 72.0 & 19.2 & 41.6 & 25.6 \\
    \midrule
    Matrix-Game~3.5 & 35.3 & 62.0 & 24.2 & 53.1 & 28.0 & 51.3 & 68.8 & 23.3 & \third{60.1} & 25.7 \\
    GEN3C & 42.4 & 54.8 & 26.8 & 37.0 & 38.2 & 44.1 & 57.9 & 25.6 & 39.1 & 28.4 \\
    Lyra~2.0$^\dagger$ & 40.5 & 62.5 & \third{31.9} & 43.9 & 41.7 & 56.1 & 70.9 & 24.0 & 40.6 & 31.5 \\
    Lyra~2.0 (DMD)$^\dagger$ & 45.5 & 57.4 & 27.1 & \third{55.8} & 38.4 & 59.7 & 70.6 & 28.3 & 58.2 & 26.7 \\
    \midrule
    WorldDirector$^\dagger$ & 37.2 & 53.3 & 24.7 & 46.3 & 49.3 & 37.1 & 57.7 & \third{38.1} & 50.6 & 33.9 \\
    Code World Model & 48.0 & \third{64.9} & 19.6 & 46.4 & \second{58.1} & 65.8 & 70.6 & 25.1 & 52.7 & \second{42.4} \\
    Code Video Model$^\S$ & 34.3 & 63.1 & 16.9 & 37.4 & 41.0 & 29.2 & \third{74.8} & 15.0 & 49.2 & 35.7 \\
    AgentGarten$^\dagger$ & \second{72.3} & \first{72.0} & \second{34.7} & \second{59.3} & \third{52.6} & \second{84.3} & \first{78.1} & \second{43.4} & \second{73.7} & \third{38.1} \\
    \midrule
    \method & \first{78.2} & \second{69.7} & \first{91.5} & \first{85.8} & \first{86.0} & \first{86.8} & \second{76.2} & \first{94.1} & \first{91.5} & \first{83.1} \\
    \bottomrule
  \end{tabular}
\vspace{-8pt}
\end{table}

%% file: sec/5_conclusion.tex
\section{Conclusion}
\label{sec:conclusion}
We present \method, a real-time video renderer that stores what it generates on the mesh it renders.
As it addresses memory by surface, \method more than doubles the best baseline's object permanence and nearly doubles cross-agent consistency, with memory that grows only with the surfaces seen, while rendering 720p video at 40~FPS.
Although a coarse mesh suffices, \method assumes that it is accurate, and handling noisy geometry reconstructed from images~\cite{burzio2026dejaview} would be a valuable next step.
We discuss limitations in \cref{sec:supp-limitations}.

\section*{Acknowledgments}
We thank Rafa{\l} Karp, Gene Liu, and the Isekai3D team~\cite{karp2026isekai3d} for developing and supporting the Isekai3D synthetic data generation platform.

\FloatBarrier

%% file: sec/X_suppl.tex
\clearpage
\makeatletter\setlength{\@fptop}{0pt}\setlength{\@dblfptop}{0pt}\makeatother%
\setcounter{table}{0}
\setcounter{figure}{0}
\setcounter{equation}{0}
\renewcommand{\thefigure}{A\arabic{figure}}
\renewcommand{\thetable}{A\arabic{table}}
\renewcommand{\theequation}{A\arabic{equation}}
\renewcommand{\theHequation}{A\arabic{equation}}
\appendix
\maketitlesupplementary

\etocdepthtag.toc{supp}
\etocsettagdepth{main}{none}
\etocsettagdepth{supp}{subsection}
\etocsettocstyle{\noindent\textbf{\large Contents}\par\vspace{0.4em}}{\vspace{1.2em}}
{\renewcommand*\numberline[1]{#1\hskip 1em}\tableofcontents}%

\input{sec/supp/A_supplementary_video}
\input{sec/supp/B_diagnostics}

\input{sec/supp/C_geometry_conditioning}
\input{sec/supp/D_surface_memory}
\input{sec/supp/E_training}
\input{sec/supp/F_fastvae}
\input{sec/supp/G_evaluation}
\input{sec/supp/H_limitations}
\input{sec/supp/I_societal_impact}
\input{sec/supp/J_qualitative}

%% file: sec/supp/A_supplementary_video.tex
\section{Videos}
\label{sec:supp-video}
We recommend watching our videos on our project page at \url{https://research.nvidia.com/labs/sil/projects/m2r/}.
The page shows single-agent and multi-agent rollouts of \method across diverse meshes, compares it to AgentGarten~\cite{chi2026agentgarten} and Code World Model~\cite{chen2026codeworld} on reopened drawers, moved objects, and multiple agents, and includes an 8{,}001-frame rollout (4\,min 27\,s).

%% file: sec/supp/B_diagnostics.tex
\section{Additional Diagnostics}
\label{sec:supp-ablations}

\paragraph{Memory without training.}
To test whether the causal renderer benefits from the memory without any training, we evaluate it before memory distillation (upper rows of \cref{tab:supp-compression}).
The memory already raises content persistence from 84.45 to 88.48, whereas widening the attention window from 8 to 64 frames only reaches 86.19 and renders slower (35.2 vs.\ 42.9~FPS).
On HSSD, it also raises revisit from 74.33 to 81.98 (\cref{tab:diagnostics,tab:supp-kept}).

\paragraph{Compression.}
We measure the impact of compression by varying the number of principal components kept per token and the precision at which we store them (\cref{tab:supp-compression}).
Content persistence improves with the number of components and saturates at 8{,}192.
Storing these components at 8-bit precision yields a memory that is as consistent as the uncompressed one, both before and after memory distillation, with tokens that are 18 times smaller.
Compression and decompression have a negligible impact on the frame rate.

\begin{table}[t]
  \centering
  \caption{\textbf{Compression.}
  We vary the number of principal components kept per token and their precision on ScanNet++, before and after memory distillation (trained, 16 GPUs).
  Store reports the peak size of the 0.485~GiB temporal cache plus the stored tokens, which shrink 18$\times$ on their own (3.70 to 0.21~GiB).
  Our compressed memory is as consistent as uncompressed keys and values (Raw K/V) at a fraction of the storage.}
  \label{tab:supp-compression}
  \footnotesize
  \setlength{\tabcolsep}{2.5pt}
  \newcommand{\hcell}[1]{\begin{tabular}[c]{@{}c@{}}#1\end{tabular}}
  \begin{tabular*}{\linewidth}{@{\extracolsep{\fill}}lccccc@{}}
    \toprule
    Memory & \hcell{Control\\Align.} & \hcell{Visual\\Quality} & \hcell{Content\\Persist.} & \hcell{Store\\(GiB)$\downarrow$} & FPS$\uparrow$ \\
    \midrule
    None (8f window) & 95.07 & 72.33 & 84.45 & 0.485 & 51.1 \\
    None (64f window) & 94.42 & 71.43 & 86.19 & 2.177 & 35.2 \\
    \midrule
    Raw K/V & 95.74 & 72.50 & 88.48 & 4.190 & 42.9 \\
    PCA 512 & 95.41 & 72.25 & 85.83 & 0.511 & 40.9 \\
    PCA 1k & 95.57 & 72.45 & 87.65 & 0.537 & 40.7 \\
    PCA 2k & 95.32 & 72.43 & 88.19 & 0.589 & 41.0 \\
    PCA 4k & 95.48 & 72.40 & 88.65 & 0.691 & 40.7 \\
    PCA 8k & 95.71 & 72.50 & 88.82 & 0.897 & 41.0 \\
    PCA 16k & 95.22 & 72.52 & 88.84 & 1.309 & 40.6 \\
    PCA 8k, FP8 & 95.57 & 72.50 & 88.64 & 0.691 & 41.0 \\
    \midrule
    Raw K/V (trained) & 95.90 & 72.62 & 89.74 & 4.190 & 42.9 \\
    \method (trained) & 95.86 & 72.68 & 89.83 & 0.691 & 41.1 \\
    \bottomrule
  \end{tabular*}
\end{table}

\paragraph{Observations per cell.}
To validate retaining both the first and the latest observation of each cell, we compare to variants that retain only one of the two (\cref{tab:supp-kept}).
Retaining both is more consistent than either alone, as the first observation preserves the initial appearance of a surface and the latest its most recent one.
Averaging the two instead of attending to each degrades consistency.

\begin{table}[t]
  \centering
  \caption{\textbf{Observations per cell.}
  We vary which observations each cell retains, before memory distillation.
  Retaining both the first and the latest observation is more consistent than retaining either alone or averaging the two.
  The frame rates differ within measurement noise.
  }
  \label{tab:supp-kept}
  \footnotesize
  \setlength{\tabcolsep}{0pt}%
  \newcommand{\hcell}[1]{\begin{tabular}[c]{@{}c@{}}#1\end{tabular}}
  \begin{tabular*}{\linewidth}{@{\extracolsep{\fill}}l>{\centering\arraybackslash}p{37pt}@{\extracolsep{0pt}}*{3}{>{\centering\arraybackslash}p{37pt}}}
    \toprule
    & \multicolumn{1}{c}{\makebox[0pt]{ScanNet++}} & HSSD & & \\
    \cmidrule(l{1.5pt}r{1.5pt}){2-2}\cmidrule(l{1.5pt}r{1.5pt}){3-3}
    Observations & \hcell{Content\\Persist.} & Revisit & \hcell{Store\\(GiB)$\downarrow$} & FPS$\uparrow$ \\
    \midrule
    First only & \third{84.66} & \second{78.90} & \second{0.625} & 40.8 \\
    Latest only & \second{88.34} & 75.83 & \first{0.617} & 41.3 \\
    First + latest & \first{88.64} & \first{81.98} & \third{0.691} & 41.0 \\
    Mean of first + latest & 84.65 & \third{76.35} & \third{0.691} & 39.6 \\
    \bottomrule
  \end{tabular*}
\end{table}

\begin{figure}[t]
  \centering
  \includegraphics[width=\linewidth]{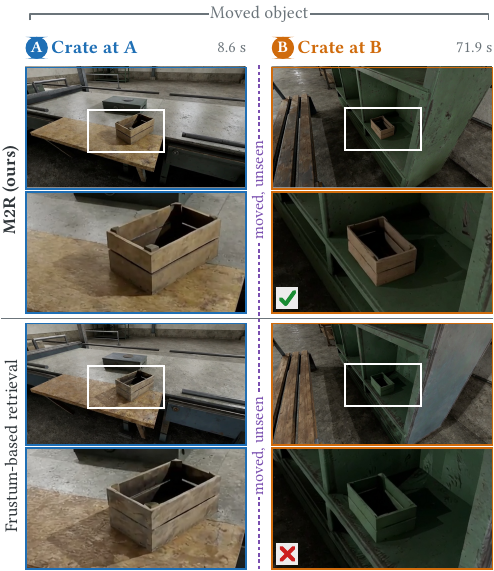}
  \caption{\textbf{Moved object under frustum retrieval.}
  A crate is seen at A and moved to B while the camera looks away.
  As the frames that show the crate at A share no view with B, frustum retrieval invents the crate's appearance there and draws it green.
  \method keeps the crate's tokens in cells that move with the crate (\cref{eq:cell}), and thus draws the wooden crate at B.}
  \label{fig:supp-frustum}
\end{figure}

\paragraph{Memory structure.}
We ablate what the memory stores by comparing it to variants that instead store colors on the same surface cells or retrieve past latent frames by view frustum~\cite{xiao2025worldmem,yu2025contextasmemory}.
We summarize our results in \cref{tab:supp-structure}.
In the color variants, each generated chunk writes RGB colors to the same surface cells.
For each subsequent chunk, we re-render the stored colors (first, latest, or both) from the current camera, encode them with FastVAE, and pass them as retrieved tokens at $t-8$ under the same 16-GPU memory distillation.
Storing keys and values is more consistent than storing colors on both content persistence and revisit, and renders faster, as colors are a lossy copy of what the renderer generated and must be re-rendered and re-encoded for every chunk.
As a retrieved frame shows the scene as it was, retrieving past frames is less consistent in dynamic scenes (\cref{fig:supp-frustum}, 82.93 vs.\ 86.78 on revisit and 80.76 vs.\ 85.98 on moved objects), although it suffices on static scenes such as ScanNet++.
On ScanNet++, every trajectory returns to its exact initial pose, and most intermediate revisits come from the same or nearby viewpoints.
Our memory is both the fastest and the most consistent on revisits and moved objects in dynamic scenes.

\begin{table}[t]
  \centering
  \caption{\textbf{Memory structure.}
  We compare our memory to variants that retrieve latent frames by view frustum or store colors (RGB) on the same surface cells, each after the same memory distillation (16 GPUs).
  Our memory is the fastest and the most consistent on revisits and moved objects in dynamic scenes.
  }
  \label{tab:supp-structure}
  \footnotesize
  \setlength{\tabcolsep}{0pt}
  \newcommand{\hcell}[1]{\begin{tabular}[c]{@{}c@{}}#1\end{tabular}}
  \begin{tabular*}{\linewidth}{@{\extracolsep{\fill}}l>{\centering\arraybackslash}p{37pt}@{\extracolsep{0pt}}*{3}{>{\centering\arraybackslash}p{37pt}}}
    \toprule
    & \multicolumn{1}{c}{\makebox[0pt]{ScanNet++}} & \multicolumn{2}{c}{HSSD} & \\
    \cmidrule(l{1.5pt}r{1.5pt}){2-2}\cmidrule(l{1.5pt}r{1.5pt}){3-4}
    Memory & \hcell{Content\\Persist.} & Revisit & \hcell{Moved\\Objects} & FPS$\uparrow$ \\
    \midrule
    Latent frames (frustum) & 91.83 & 82.93 & 80.76 & 15.5 \\
    RGB, first & 83.52 & 79.10 & 85.92 & 29.6 \\
    RGB, latest & 85.06 & 78.56 & 84.55 & 29.0 \\
    RGB, first + latest & 83.59 & 79.56 & 83.48 & 19.8 \\
    \midrule
    K/V (\method) & 89.83 & 86.78 & 85.98 & 41.1 \\
    \bottomrule
  \end{tabular*}
\end{table}

%% file: sec/supp/C_geometry_conditioning.tex
\section{Geometry Conditioning}
\label{sec:supp-rendering}
\begin{figure*}[tp]
  \centering
  \includegraphics[width=\linewidth]{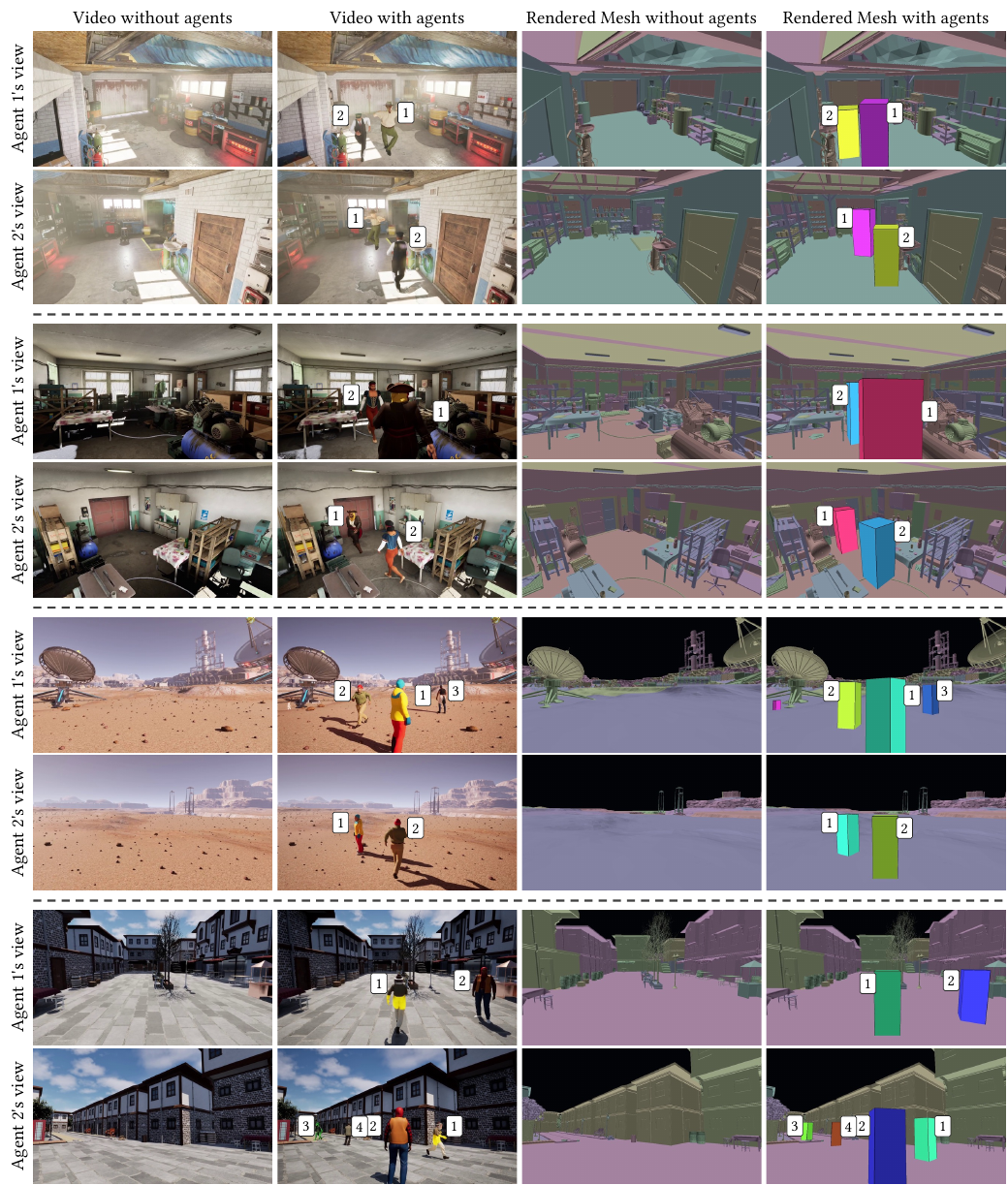}
  \caption{\textbf{Proxy video.}
  We show training videos (\textbf{left}) and their proxy videos (\textbf{right}) for four episodes, each rendered without and with agents.
  The proxy videos carry the geometry of each view but none of its appearance, and show each agent as a box (the numbers mark the same agent in both views).}
  \label{fig:supp-proxy}
\end{figure*}

\paragraph{Proxy rendering.}
At every video frame, we rasterize the current mesh from each agent's camera into a proxy video $\mathbf{R}$ (\cref{fig:supp-proxy}).
We use flat shading under a fixed directional light and assign each asset one arbitrary color, which separates objects but carries no information about their appearance.
Pixels without geometry, such as the sky, remain black.
We encode $\mathbf{R}$ with the same autoencoder $\mathcal{E}$ as the video and add it to the video tokens through a zero-initialized projection $W_m$, such that each token $i$ receives the geometry it should depict:
\begin{equation}
  \mathbf{h}_i = W_v\,\mathcal{P}(\mathbf{z}_{\tau})_i + W_m\,\mathcal{P}\big(\mathcal{E}(\mathbf{R})\big)_i,
  \label{eq:supp-geometry-conditioning}
\end{equation}
where $\mathbf{h}_i$ is the embedding of token $i$, $\mathcal{P}$ patchifies latents, and $W_v$ is the pretrained projection of the noisy latents $\mathbf{z}_{\tau}$.

\paragraph{Online rasterization.}
At inference and during memory distillation, we rasterize the mesh on the GPU with \mbox{nvdiffrast}~\cite{laine2020modular}, in the same process as the renderer.
Besides the proxy video, the rasterizer provides the depth, object, and surface normal behind each pixel, which the surface memory uses to assign pixels to cells (\cref{sec:supp-memory}).
To limit the rasterizer's GPU memory during training, we cap the scene geometry at 4\,GiB (about 70 million triangles) and simplify larger training scenes offline to 1\% of the triangles of each asset with quadric error metrics~\cite{garland1997surface}.

\paragraph{Agents.}
We draw each agent into the proxy video as a bounding box (\cref{fig:supp-proxy}), from which the renderer generates the agent's body and motion.
The hue of a box is random per episode and independent of the agent's identity, while its sides differ in brightness to show the agent's heading.
All human characters share a box of $68\times72\times184$\,cm, and each non-human character has a box that fits its asset.
A third-person view shows the box of the agent that it follows and the boxes of all other agents in view, whereas a first-person view shows only the boxes of the other agents.

\paragraph{Keyboard controls.}
The renderer does not require keyboard controls, which serve as a hint for how to animate an agent.
During training, each view receives the keys held at each video frame by the agent that it follows (W, A, S, D, space, shift, and ctrl), or a flag that marks them as unknown if the clip has no agents.
We embed the keys of the four video frames that each latent frame spans with a two-layer SiLU MLP whose last layer is zero-initialized, and add the embedding to every token of that latent frame.
At evaluation, including episodes with agents, keys are unavailable, so instead of synthesizing them from the bounding boxes, we simply mark them as unknown, and the renderer still animates each agent naturally from its box alone.

%% file: sec/supp/D_surface_memory.tex
\section{Surface Memory}
\label{sec:supp-memory}
\paragraph{Cells.}
We extend the cells of \cref{eq:cell} so that every pixel of a view maps to a cell, and what the renderer generates thus persists on distant surfaces, on thin structures, on agents, and where the mesh has no geometry (\cref{tab:supp-memory-cells}).
As a pixel covers a larger patch of a surface the farther away the surface is, we keep cells at four sizes $s_\ell$, from 3\,cm to 1.92\,m, and let each view write to those that match what its pixels cover.
As a cubic cell can hold surfaces that face different ways, such as the two sides of a thin wall, we also split each cell by the direction $d(p)$ that its surface faces, and \cref{eq:cell} becomes
\begin{equation}
  c_\ell(p) = \Big(o(p),\; \big\lfloor \mathbf{T}_{o(p)}(t)\,\mathbf{x}(p) / s_\ell \big\rfloor,\; d(p)\Big),
  \label{eq:supp-cell}
\end{equation}
where $d(p)$ is one of six directions, the dominant axis and sign of the surface normal in the object's coordinates.
An agent's box has cells of its own in the coordinates of the box, such that the agent carries its appearance with it (\cref{tab:multiagent}).

Pixels without a surface behind them have no cell under \cref{eq:supp-cell}.
Where such a pixel is enclosed by geometry, as in the holes of scanned meshes (\cref{fig:supp-mesh-repair}), we interpolate its depth from its surroundings and use the cells of the resulting point.
Otherwise, as for the sky, we define its cells by viewing direction per 2\,m camera-position cell, so they also remain bounded by the explored region of the scene.

\begin{table}[t]%
  \centering
  \caption{\textbf{Cells for every kind of pixel.}
  We list where the cells are defined and their sizes.
  Every pixel of a view thus maps to a cell, so that appearance persists on surfaces, on agents, and where the mesh has no geometry.}
  \label{tab:supp-memory-cells}
  \footnotesize
  \begin{tabular}{@{}lp{0.42\linewidth}l@{}}
    \toprule
    Pixel on & Cells defined in & Cell sizes \\
    \midrule
    Mesh surface & Object coordinates, split by six surface directions & 3, 12, 48\,cm, 1.92\,m \\
    Agent box & Box coordinates & 5, 20\,cm \\
    Hole & World coordinates, at an interpolated depth & 12, 48\,cm \\
    Sky & Viewing direction, per 2\,m cell of the camera position & 8$^\circ$, 32$^\circ$ \\
    \bottomrule
  \end{tabular}
\end{table}

\paragraph{Writing.}
Each view writes only to the cells whose size matches the scale at which it sees a surface (fine cells for a close-up and coarse cells for a distant view).
We measure this scale by the footprint $f(p)$ of each pixel $p$, the width of the patch of surface that it covers.
As a token $j$ spans many pixels, it covers a larger patch of width $F(j)$.
On mesh surfaces, a pixel $p$ of token $j$ writes to the cells of every size from $f(p)$ to $F(j)$,
\begin{equation}
  \mathcal{W}(p) = \{\ell : f(p) \le s_\ell \le F(j)\}, \quad p \in \Omega_j.
  \label{eq:supp-write-levels}
\end{equation}
If no size lies in this range, the pixel ignores the upper bound and writes to the smallest $s_\ell \ge f(p)$.
Agent boxes, holes, and the sky each have only two cell sizes (\cref{tab:supp-memory-cells}), and their pixels write to both.
In practice, we rasterize the mesh at half the video resolution for efficiency, as cells and tokens are both far coarser than a pixel.
A token then spans $16\times16$ pixels, and we set $F(j)$ to 16 times the median footprint of its pixels.

\paragraph{Storage.}
We compress each stored token 18-fold without degrading consistency (\cref{tab:supp-compression}).
The keys and values of a token form a vector $\mathbf{v}$ of $2 \times 36 \times 8 \times 128 = 73{,}728$ numbers, over 36 layers with 8 heads of 128 dimensions.
We fit a principal component basis to these vectors once, on rollouts of the causal renderer before memory distillation.
We then project each vector onto this basis and store its coefficients at 8-bit precision (FP8),
\begin{equation}
  \mathbf{q} = \mathrm{FP8}\Big(\boldsymbol{\Lambda}^{-1}\,\mathbf{U}^{\top}\mathbf{S}^{-1}(\mathbf{v} - \boldsymbol{\mu})\Big),
  \label{eq:supp-pca}
\end{equation}
where $\boldsymbol{\mu}$ is the mean of $\mathbf{v}$, the diagonal matrix $\mathbf{S}$ holds its standard deviation per head, the columns of $\mathbf{U}$ are the 8{,}192 leading principal components, and the diagonal matrix $\boldsymbol{\Lambda}$ scales each coefficient to unit variance.
When a token is read, we invert these steps, $\hat{\mathbf{v}} = \boldsymbol{\mu} + \mathbf{S}\,\mathbf{U}\boldsymbol{\Lambda}\,\mathbf{q}$.
Each stored token takes 8{,}192 bytes instead of 147{,}456 at 16-bit precision.
On the 46 ScanNet++ trajectories, the stored tokens thus take at most 0.21~GiB instead of 3.70~GiB.
Over the 8{,}001-frame rollout of \cref{fig:persistence}b, they take 5.95~GiB instead of 107~GiB.
Even a rollout that kept every token it generates would add 1.7~GiB per 1{,}000 frames at $704\times1280$ instead of 30~GiB.

\paragraph{Reading.}
As in writing, each pixel reads at the scale at which it sees a surface.
On mesh surfaces, a pixel $p$ reads the finest cell that holds an observation and is at least as large as its footprint ($s_\ell \ge f(p)$).
Without this bound, a distant view would read the fine cells of an earlier close-up instead of the cells at its own scale.
For agent boxes, holes, and the sky, we omit this bound, and a pixel reads the finest of its two cells that holds an observation.
Each token only retrieves past tokens from earlier visits to a surface.
We thus exclude from the selection of \cref{eq:read-vote} the observations made since the surface last came into view, as well as those whose tokens are still in the temporal cache, as the reading token attends to them directly.
The selection with the latest observations also skips the chunk already selected with the first observations, such that the two retrieved tokens come from different chunks.

\begin{figure}[t]
  \centering
  \includegraphics[width=\linewidth]{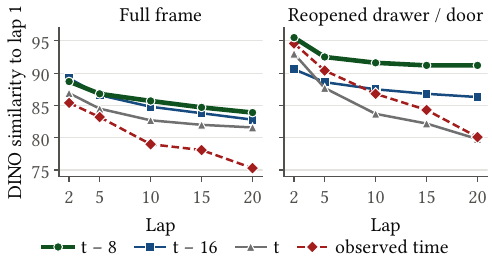}
  \caption{\textbf{Time position of retrieved tokens.}
  We compare four positions: 8 or 16 latents before the reading token ($t-8$, $t-16$), its own time ($t$), and the time at which the retrieved tokens were observed.
  For this figure, we change the position at inference only and plot an earlier renderer trained with retrieved tokens at $t$ (our final renderer is trained with them at $t-8$).
  We plot how similar each lap is to the first, over the full frame (\textbf{left}) and inside reopened drawers and doors (\textbf{right}), for 20 laps of 13 HSSD loops (about 340 frames per lap, 7 minutes in total).
  Whereas \cref{fig:persistence}a measures consistency on a single revisit at varying time intervals, here each lap is compared to the first across 20 repeated revisits of the same loop.
  Our choice ($t-8$) is the most consistent over long rollouts, while the observed time degrades the most, as its distance to the reading token keeps growing.
  }
  \label{fig:supp-read-time}
\end{figure}

\paragraph{Memory conditioning.}
In every layer, each token attends to the tokens retrieved for its chunk, both in the denoising steps and when the generated chunk is re-encoded for the temporal cache.
We decompress the retrieved tokens into keys and values before attention, and never add them to this cache.
As keys are stored without positional encoding~\cite{xiao2024attentionsinks,liu2025rollingforcing,wu2026addressable}, we apply the backbone's rotary encoding~\cite{nvidia2026cosmos3} when they are read.
We give each retrieved token the spatial position of the reading token and a time that is a fixed 8 latents (32 frames) earlier, regardless of when it was observed.
A token retrieved by token $i$ at latent time $t_i$ and spatial position $(h_i, w_i)$ is thus placed at
\begin{equation}
  (t_i - 8,\; h_i,\; w_i).
  \label{eq:supp-read-position}
\end{equation}
With chunks of 2 latents and a cache of one chunk, the current chunk sits at $\{t-1, t\}$, the previous chunk at $\{t-3, t-2\}$, and the re-anchored initial chunk at $t-4$, such that the retrieved tokens at $\{t-9, t-8\}$ lie beyond the temporal cache.
We compare this placement to alternatives in \cref{fig:supp-read-time}.
Larger offsets (e.g., 16 or 32 latents) also work, while placing retrieved tokens at the time they were observed is less consistent over long rollouts.

\paragraph{Multiple agents.}
Several agents can share the surface memory although we train with a single view at a time, as the memory indexes tokens by the surfaces they depict rather than by the view that generated them (\cref{tab:multiagent}).
At inference, all agents generate a chunk before any of them writes, and their observations then enter the memory together, ordered by agent index where several agents write to the same cell.
The result thus does not depend on the order in which the agents run.
Each agent runs on its own GPU.
Each agent reads the observations of the other agents under the same rules as its own, including those from their first frames, which are not in its temporal cache.
It skips the observations that another agent made in the previous chunk or since the surface last came into its own view.
Two agents that look at the same surface would otherwise read each other's latest frames at every chunk, causing temporal inconsistencies.

Agents also keep their appearance from one meeting to the next (\cref{tab:multiagent}).
In third-person videos, each agent starts from a first frame that shows it only from behind, so another view that first sees its front has to guess its appearance.
As a written guess would stay in the memory for the rest of the video, an agent's box is only written by its own view and by the given first frames of other views, which are not generated.
A token on the box in turn attends only to the tokens retrieved for itself, as those retrieved for the rest of its chunk would mix in the appearance of other agents.
It also attends to reference tokens of that agent, taken from the agent's own view (its first frame and its latest chunk that shows the agent) and from the given first frames of other views that show it.
We empirically place these references 48 latents before the reading token (16 or 32 also work), apart from the retrieved tokens at $t-8$, and add $\ln 4$ to their attention logits, which weights them four-fold so that the renderer draws unseen sides of an agent consistently with its observed views (\cref{sec:supp-limitations}).

%% file: sec/supp/E_training.tex
\begin{figure}[t]
  \centering
  \includegraphics[width=\linewidth]{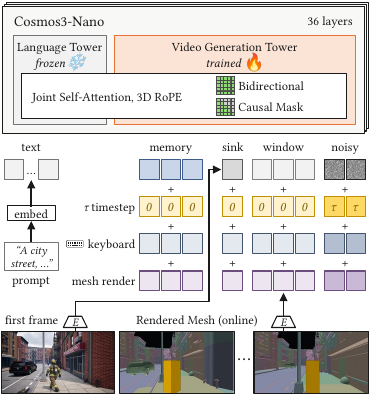}
  \caption{\textbf{Renderer architecture.}
  We build our renderer on Cosmos3-Nano and train only its video tower.
  Each video token receives the embeddings of its timestep, keyboard controls, and mesh render, and the text tokens are a fixed prefix.
  The same network runs with bidirectional attention in the first stage, with a causal mask in teacher forcing and Causal CD, and with an autoregressive key-value cache in Self Forcing and memory distillation.
  In these two stages, only the noisy chunk runs through the network, while the sink and window tokens (and the memory tokens in memory distillation) enter as a key-value cache that we commit from a clean pass.
  The cached keys of the sink and the window already carry their timestep and rotary position, whereas memory tokens are stored without rotary position, which we apply whenever they are read (\cref{sec:supp-memory}).}
  \label{fig:supp-architecture}
\end{figure}

\section{Training and Inference}
\label{sec:supp-optimization}
\label{sec:supp-data}
\label{sec:supp-inference}
\paragraph{Data.}
Although we fine-tune Cosmos3-Nano~\cite{nvidia2026cosmos3} (\cref{fig:supp-architecture}) only on synthetic episodes rendered in Unreal Engine with Isekai3D~\cite{karp2026isekai3d}, \method transfers zero-shot to every dataset that we evaluate on (\cref{sec:supp-evaluation}), including real indoor scans and worlds authored by coding agents.
The training set has about 200{,}000 episodes of 60\,s at 30~FPS and $720\times1280$, which span 223 environments.
Each episode has one to eight agents drawn from 367 human and 34 non-human characters, and every agent is followed by its own camera, which gives about 850{,}000 single-view videos.
We render non-human characters only in single-agent episodes.
We render each video a second time along the same camera path without agents, and each render comes with its proxy video (\cref{fig:supp-proxy}).
Until memory distillation, half of the training clips show agents and the other half do not.
We train on one view at a time, on clips of 129 to 513 frames at $480\times832$ (480p) or $704\times1280$ (720p).
Despite never training on two views of a world together or on more than 513 frames, \method nearly doubles the best cross-agent consistency among the baselines (\cref{tab:multiagent}) on episodes of two to four times that length (\cref{sec:supp-demos}).

\paragraph{Stages and hyperparameters.}
We train \method in five stages on 64 GB200 GPUs (\cref{tab:supp-training}), starting each stage from the exponential moving average (EMA) of the weights of the previous one.
We first train a bidirectional model for 23{,}000 iterations on clips of 129, 257, and then 513 frames at both resolutions, using AdamW~\cite{loshchilov2019adamw} with a learning rate of $1\times10^{-5}$.
We then make it causal with 2{,}000 iterations of teacher forcing at the same learning rate, and distill it into four steps with 500 iterations of Causal CD~\cite{zhao2026causal} at $2\times10^{-6}$.
We next train it on its own rollouts with 2{,}000 iterations of Self Forcing, using learning rates of $2\times10^{-6}$ for the generator and $4\times10^{-7}$ for the fake score function.
Throughout causal training, we draw each chunk's sliding window uniformly from one to 15 chunks or the entire past.
Finally, we teach the renderer to use its memory with only 300 iterations of memory distillation on 513-frame clips at $704\times1280$, with $\lambda = 0.1$.

\begin{table}[t]
  \centering
  \caption{\textbf{Training stages.} We train \method in the order listed.
  Batch lists the clips per iteration at 480p\,/\,720p.
  We draw both resolutions equally often for the first 19{,}000 iterations of the bidirectional stage, and 480p for one iteration in five for the rest of that stage and throughout the three causal stages.}
  \label{tab:supp-training}
  \footnotesize
  \setlength{\tabcolsep}{4pt}
  \begin{tabular*}{\linewidth}{@{\extracolsep{\fill}}lcccc@{}}
    \toprule
    Stage & Frames & Resolution & Iterations & Batch \\
    \midrule
    Bidirectional & 129 & 480p, 720p & 13{,}000 & 576\,/\,256 \\
     & 257 & 480p, 720p & 2{,}000 & 256\,/\,128 \\
     & 513 & 480p, 720p & 8{,}000 & 256\,/\,128 \\
    \midrule
    Teacher forcing & 513 & 480p, 720p & 2{,}000 & 64 \\
    Causal CD~\cite{zhao2026causal} & 513 & 480p, 720p & 500 & 64 \\
    Self Forcing & 513 & 480p, 720p & 2{,}000 & 64 \\
    \midrule
    Memory distillation & 513 & 720p & 300 & 64 \\
    \bottomrule
  \end{tabular*}
\end{table}

\renewcommand{\thealgorithm}{A\arabic{algorithm}}
\begin{algorithm}[t]
  \caption{Memory distillation (one iteration)}
  \label{alg:supp-memory-distillation}
  \footnotesize
  \begin{algorithmic}[1]
    \Require Renderer $G_\theta$, its frozen copy $G_{\bar\theta}$, a clip of $N$ chunks
    \State Empty the surface memory $\mathcal{M}$ and pick the exit step $k$
    \For{$n = 1, \dots, N$}
      \State $\mathcal{R}_n \gets \textsc{Read}(\mathcal{M})$ \Comment{retrieved tokens}
      \State $\hat{\mathbf{z}}^{S}_n \gets G_\theta$ in $k$ steps, given the cache and $\mathcal{R}_n$
      \If{the clip is a palindrome}
        \State $\hat{\mathbf{z}}^{T}_n \gets G_{\bar\theta}$, given all earlier chunks
      \EndIf
      \State Re-encode $\hat{\mathbf{z}}^{S}_n$, update the cache, $\textsc{Write}(\mathcal{M})$
    \EndFor
    \If{generator iteration (one in six)}
      \State Update $\theta$ with $\mathcal{L}_{\mathrm{DMD}} + \lambda\,\mathcal{L}_{\mathrm{SD}}$
    \Else
      \State Update the fake score function
    \EndIf
  \end{algorithmic}
\end{algorithm}

\begin{figure}[t]
  \centering
  \includegraphics[width=0.8\linewidth]{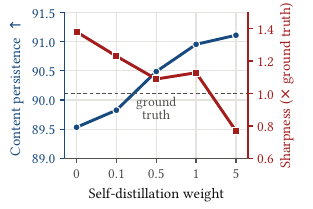}
  \caption{\textbf{Self-distillation weight.}
  We vary the weight $\lambda$ of the self-distillation loss in memory distillation, training each model on 16 GPUs.
  We plot content persistence on ScanNet++ and sharpness relative to the ground truth on 513-frame revisiting trajectories in a world we build with a coding agent.
  A larger weight trades sharpness for consistency, and \method uses 0.1.}
  \label{fig:supp-sd-weight}
\end{figure}

\paragraph{Memory distillation.}
In each iteration of memory distillation (\cref{alg:supp-memory-distillation}), the renderer rolls out all chunks of a clip with a temporal cache of one chunk, reading from and writing to the surface memory as at inference (\cref{sec:supp-memory}).
In 70\% of the iterations, the clip is a palindrome with one, two, or four round trips, which we build from agent-free clips only.
In the rest, it is a clip with agents, played forward.
We apply $\mathcal{L}_{\mathrm{DMD}}$ to every clip, with a teacher and a fake score function that see the whole clip and have no memory.
We apply $\mathcal{L}_{\mathrm{SD}}$ to palindromes only and vary its weight $\lambda$ in \cref{fig:supp-sd-weight}.
In the ablation without revisiting rollouts (\cref{tab:diagnostics}), we apply $\mathcal{L}_{\mathrm{SD}}$ to forward agent-free clips using the same frozen teacher.
A larger $\lambda$ improves consistency but softens textures.
Because the synthetic ground-truth renders in \cref{fig:supp-sd-weight} are less vivid and less detailed than photorealistic video, we empirically choose $\lambda = 0.1$ for the best balance of visual quality and consistency across diverse scenes.

\paragraph{Sampling.}
At test time, \method renders each eight-frame chunk in two steps, at noise levels 1 and 0.625, without classifier-free guidance.
Although we distill the renderer into four steps, we find that two-step and four-step results differ negligibly, as the mesh geometry and the memory provide strong signals.
The renderer is trained on 30~FPS videos, as used in the 8{,}001-frame rollout of \cref{fig:persistence}b, whereas our quantitative benchmarks follow Orbia's 16~FPS convention (\cref{sec:supp-evaluation}).

%% file: sec/supp/F_fastvae.tex
\section{FastVAE}
\label{sec:supp-fastvae}
\begin{figure}[t]
  \centering
  \includegraphics[width=\linewidth]{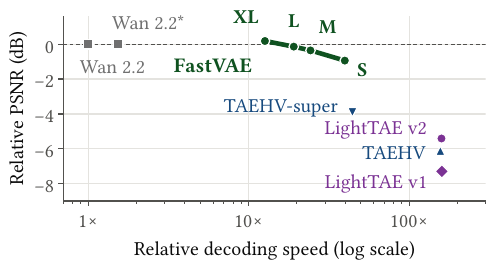}
  \caption{\textbf{Quality against speed.}
  We plot the PSNR of each decoder against its decoding speed, both relative to the Wan~2.2 decoder in eager mode and averaged over the three sizes of \cref{tab:supp-fastvae-quality}.
  The decoders run on the latents of the Wan~2.2 encoder, and we measure speed on a GB200.
  Wan~2.2 is the original decoder as it is commonly used, and Wan~2.2$^*$ is the same decoder optimized with \texttt{torch.compile} and CUDA graphs.
  All other decoders use the same optimization.}
  \label{fig:supp-fastvae-pareto}
\end{figure}

\begin{figure*}[t]
  \centering
  \includegraphics[width=\linewidth]{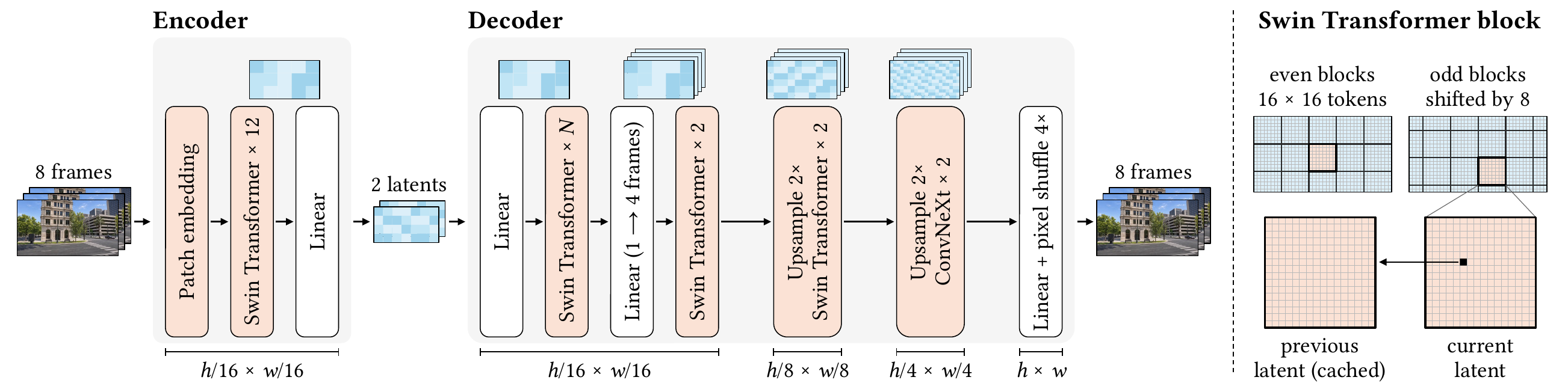}
  \caption{\textbf{FastVAE.}
  The encoder maps each chunk of eight frames to two latents in the latent space of the Wan~2.2 VAE, and the decoder maps them back to frames.
  Swin Transformer blocks attend inside shifted windows to the current and the previous latent, whose keys and values are cached, so both networks run chunk by chunk.
  \Cref{tab:supp-fastvae-sizes} lists the number of blocks $N$ and the widths of each decoder size.}
  \label{fig:supp-fastvae}
  \par\vspace{8pt}%
  \includegraphics[width=\linewidth]{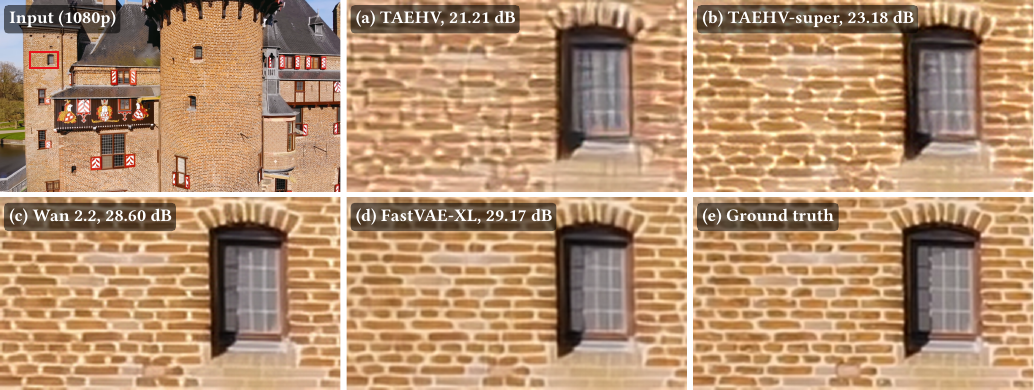}
  \caption{\textbf{Reconstruction at 1080p.}
  We decode the Wan~2.2 latents of an UltraVideo clip~\cite{xue2025ultravideo} with two tiny decoders (\textbf{a}, \textbf{b}), the Wan~2.2 decoder (\textbf{c}), and FastVAE-XL (\textbf{d}), and show the region in the red box with the PSNR of each decoder on the clip.
  The tiny decoders lose the bricks and the window bars, while FastVAE-XL is as close to the ground truth (\textbf{e}) as the Wan~2.2 decoder.
  The layout is inspired by FlashDecoder~\cite{kang2026flashdecoder}.}
  \label{fig:supp-fastvae-recon}
\end{figure*}

The Wan~2.2 VAE~\cite{wan2025wan} of the base model provides high-quality reconstructions but is slow, taking 225\,ms to decode an eight-frame chunk at $704\times1280$ on a GB200 (and still 143\,ms when compiled), more than half of the 416\,ms that \method spends on the chunk when using it.
Open-source tiny decoders for the same latents~\cite{BoerBohan2025TAEHV,lightx2v,shin2026motionstream} are far faster but degrade PSNR by 4 to 7\,dB (\cref{fig:supp-fastvae-pareto}).
FlashDecoder~\cite{kang2026flashdecoder} instead replaces the decoder with a streaming transformer but keeps the original encoder, which \method runs on the proxy video of every chunk (\cref{sec:supp-rendering}).
We thus build upon FlashDecoder and learn the encoder as well, allowing FastVAE to replace the entire Wan~2.2 VAE without changing its latent space (\cref{fig:supp-fastvae}).
FastVAE matches the quality of the original while running an order of magnitude faster (\cref{tab:supp-fastvae-quality}), more than doubling the frame rate of \method (19.2 to 41.1~FPS) without degrading consistency (\cref{tab:diagnostics}).

\paragraph{Architecture.}
As in FlashDecoder, our decoder is a streaming transformer that flattens each latent frame of size $H' \times W'$ into $H'W'$ tokens, where each latent frame attends to itself and the previous one through cached keys and values (\cref{fig:supp-fastvae}).
However, FlashDecoder attends over the whole frame, at a cost that grows quadratically with resolution.
We instead use Swin Transformer blocks~\cite{liu2021swin}, which attend within $16\times16$ windows (shifted by 8 tokens in every other block) at a cost that grows only linearly.
With full-frame attention, the same decoder takes 37\% longer at $704\times1280$ and 99\% longer at $1088\times1920$ on a GB200, with no noticeable gain in quality.
To map tokens to pixels, FlashDecoder uses a single $16\times$ pixel shuffle, whereas we upsample progressively, by $2\times$ with Swin Transformer blocks, by $2\times$ with ConvNeXt blocks~\cite{liu2022convnext}, and by $4\times$ with a pixel shuffle.
We build the encoder from the same blocks over spatiotemporal patches of $4\times16\times16$ pixels, such that it encodes a chunk in 3.5\,ms instead of the 63.5\,ms of the Wan~2.2 encoder.
The decoder comes in four sizes (\cref{tab:supp-fastvae-sizes}), which trade speed for quality (\cref{fig:supp-fastvae-pareto}).
Throughout the paper, \method uses FastVAE-XL, for both its scores and its frame rate.

\begin{table}[t]
  \centering
  \caption{\textbf{FastVAE decoder sizes.}
  $N$ and $d$ are the number and width of the Swin Transformer blocks at latent resolution, $d_\mathrm{r}$ is the width of the two blocks that refine the four frames of each latent, and $d_\mathrm{up}$ gives the widths of the two upsampling stages.
  All sizes use the same encoder architecture, 12 blocks of width 768 with 106M parameters.}
  \label{tab:supp-fastvae-sizes}
  \footnotesize
  \begin{tabular*}{\linewidth}{@{\extracolsep{\fill}}lccccc@{}}
    \toprule
    Size & $N$ & $d$ & $d_\mathrm{r}$ & $d_\mathrm{up}$ & Params (M) \\
    \midrule
    S & 8 & 1024 & 256 & 128, 64 & 125.5 \\
    M & 12 & 1024 & 512 & 256, 128 & 195.3 \\
    L & 12 & 1536 & 512 & 256, 128 & 417.3 \\
    XL & 16 & 1536 & 512 & 512, 256 & 559.9 \\
    \bottomrule
  \end{tabular*}
\end{table}

\paragraph{Training.}
As FastVAE has to stay in the latent space of the Wan~2.2 VAE, we train its decoder and encoder one at a time against the frozen original, and then both end to end.
The decoder first learns to reconstruct a clip $\mathbf{x}$ from the latents of the frozen Wan~2.2 encoder, at $224\times224$ and then at 480p to 1080p, with
\begin{equation}
  \mathcal{L}_\mathrm{dec} = \|\hat{\mathbf{x}} - \mathbf{x}\|_1 + \lambda_2\,\|\hat{\mathbf{x}} - \mathbf{x}\|_2^2 + \lambda_\mathrm{p}\,\mathrm{LPIPS}(\hat{\mathbf{x}}, \mathbf{x}),
  \label{eq:supp-fastvae-loss}
\end{equation}
where $\lambda_2 = 10$ and $\lambda_\mathrm{p} = 0.5$.
We drop the adversarial loss of FlashDecoder, which had a negligible effect on our decoder.
The encoder then learns to regress the same latents with the decoder frozen, and a final end-to-end phase trains both networks to reconstruct the clip through the encoder and the decoder.
To match the general-purpose latent space of the Wan~2.2 VAE, all phases train FastVAE on a mixture of publicly released web videos~\cite{nan2025openvid,lin2024open} and synthetic episodes rendered with Isekai3D~\cite{karp2026isekai3d}, whereas the renderer itself is fine-tuned only on synthetic data (\cref{sec:supp-data}).
We optimize weight matrices with Muon~\cite{liu2025muon} and the remaining parameters with AdamW~\cite{loshchilov2019adamw}, on 64 A100 GPUs.

\begin{table*}[!t]
  \centering
  \caption{\textbf{Reconstruction quality and speed.}
  We reconstruct 25-frame clips of the 1{,}145 UltraVideo videos~\cite{xue2025ultravideo} at three resolutions, with decoders on Wan~2.2 latents (\textbf{top}) and with each model's own encoder (\textbf{bottom}).
  We report PSNR, LPIPS with VGG features~\cite{zhang2018lpips}, rFVD (the content-debiased FVD~\cite{ge2024content} on the first 16 frames of each clip), and the time per eight-frame chunk at $704\times1280$ on a GB200 with \texttt{torch.compile} and CUDA graphs.
  FastVAE-XL stays within 0.5\,dB of Wan~2.2 in PSNR at every resolution while encoding 11$\times$ and decoding 8$\times$ faster.
  Against Wan~2.2 in the common eager mode (63.5\,ms to encode and 224.8\,ms to decode), it is 18$\times$ and 13$\times$ faster.}
  \label{tab:supp-fastvae-quality}
  \footnotesize
  \setlength{\tabcolsep}{4.15pt}%
  \begin{tabular}{@{}llccccccccccccc}
    \toprule
    & & \multicolumn{2}{c}{Params (M)} & \multicolumn{2}{c}{Time (ms)$\downarrow$} & \multicolumn{3}{c}{480p ($480\times832$)} & \multicolumn{3}{c}{720p ($704\times1280$)} & \multicolumn{3}{c}{1080p ($1088\times1920$)} \\
    \cmidrule(lr){3-4}\cmidrule(lr){5-6}\cmidrule(lr){7-9}\cmidrule(lr){10-12}\cmidrule(lr){13-15}
    Encoder & Decoder & Enc. & Dec. & Enc. & Dec. & PSNR$\uparrow$ & LPIPS$\downarrow$ & rFVD$\downarrow$ & PSNR$\uparrow$ & LPIPS$\downarrow$ & rFVD$\downarrow$ & PSNR$\uparrow$ & LPIPS$\downarrow$ & rFVD$\downarrow$ \\
    \midrule
    Wan~2.2 & Wan~2.2 & 149.6 & 555.0 & 38.8 & 142.7 & \second{36.69} & \first{0.037} & \first{4.24} & \second{38.32} & \first{0.036} & \first{3.01} & \second{40.23} & \first{0.035} & \first{2.03} \\
    Wan~2.2 & TAEHV & 149.6 & 9.9 & 38.8 & 1.4 & 30.72 & 0.122 & 17.02 & 32.21 & 0.115 & 11.32 & 33.94 & 0.107 & 8.69 \\
    Wan~2.2 & TAEHV-super & 149.6 & 19.0 & 38.8 & 5.1 & 32.62 & 0.095 & 8.55 & 34.39 & 0.088 & 5.59 & 36.49 & 0.082 & 4.03 \\
    Wan~2.2 & LightTAE~v1 & 149.6 & 9.9 & 38.8 & 1.4 & 29.87 & 0.194 & 23.29 & 31.06 & 0.196 & 17.79 & 32.40 & 0.198 & 15.06 \\
    Wan~2.2 & LightTAE~v2 & 149.6 & 9.9 & 38.8 & 1.4 & 31.37 & 0.145 & 17.62 & 32.91 & 0.132 & 11.36 & 34.72 & 0.120 & 8.32 \\
    Wan~2.2 & FastVAE-S & 149.6 & 125.5 & 38.8 & 5.7 & 35.62 & 0.062 & 6.03 & 37.37 & 0.057 & 3.85 & 39.44 & 0.053 & 2.55 \\
    Wan~2.2 & FastVAE-M & 149.6 & 195.3 & 38.8 & 9.3 & 36.25 & 0.054 & 5.31 & 37.99 & 0.050 & 3.51 & 39.94 & 0.046 & 2.41 \\
    Wan~2.2 & FastVAE-L & 149.6 & 417.3 & 38.8 & 11.6 & \third{36.45} & \third{0.052} & \third{5.02} & \third{38.19} & \third{0.048} & \third{3.29} & \third{40.19} & \third{0.045} & \third{2.22} \\
    Wan~2.2 & FastVAE-XL & 149.6 & 559.9 & 38.8 & 17.4 & \first{36.77} & \second{0.048} & \second{4.77} & \first{38.50} & \second{0.045} & \second{3.15} & \first{40.53} & \second{0.041} & \second{2.06} \\
    \midrule
    TAEHV & TAEHV & 1.5 & 9.9 & 1.0 & 1.4 & 30.25 & 0.140 & 21.68 & 31.64 & 0.131 & 14.52 & 33.25 & 0.123 & 11.12 \\
    TAEHV-super & TAEHV-super & 1.5 & 19.0 & 1.0 & 5.1 & 31.85 & 0.124 & 13.32 & 33.36 & 0.116 & 8.82 & 35.11 & 0.108 & 6.83 \\
    LightTAE~v1 & LightTAE~v1 & 1.5 & 9.9 & 1.0 & 1.4 & 28.09 & 0.251 & 56.02 & 29.04 & 0.248 & 46.95 & 30.05 & 0.245 & 40.02 \\
    LightTAE~v2 & LightTAE~v2 & 1.5 & 9.9 & 1.0 & 1.4 & 28.74 & 0.215 & 52.16 & 29.76 & 0.201 & 40.32 & 30.86 & 0.188 & 33.14 \\
    FastVAE-S & FastVAE-S & 106.2 & 125.5 & 3.5 & 5.7 & 35.42 & 0.074 & 6.12 & 37.17 & 0.067 & 4.05 & 39.24 & 0.060 & 2.65 \\
    FastVAE-M & FastVAE-M & 106.2 & 195.3 & 3.5 & 9.3 & \third{35.87} & \third{0.067} & \third{5.69} & \third{37.62} & \third{0.061} & \third{3.69} & \third{39.61} & \third{0.055} & \third{2.59} \\
    FastVAE-L & FastVAE-L & 106.2 & 417.3 & 3.5 & 11.6 & \second{36.02} & \second{0.065} & \second{5.45} & \second{37.77} & \second{0.059} & \second{3.59} & \second{39.82} & \second{0.054} & \second{2.39} \\
    FastVAE-XL & FastVAE-XL & 106.2 & 559.9 & 3.5 & 17.4 & \first{36.24} & \first{0.062} & \first{5.09} & \first{37.99} & \first{0.056} & \first{3.30} & \first{40.05} & \first{0.051} & \first{2.24} \\
    \bottomrule
  \end{tabular}
\end{table*}

\paragraph{Speed and fidelity.}
Following FlashDecoder, we evaluate reconstruction on UltraVideo~\cite{xue2025ultravideo} (\cref{tab:supp-fastvae-quality,fig:supp-fastvae-pareto}).
All four sizes of FastVAE outperform the open-source tiny decoders in every metric and at every resolution, and the smallest (FastVAE-S) is already 3\,dB above the best of them at a similar decoding time (5.7 vs.\ 5.1\,ms).
FastVAE-XL exceeds the PSNR of the Wan~2.2 decoder at every resolution while decoding 8$\times$ faster than its compiled version, and stays within 0.5\,dB of the full Wan~2.2 VAE with its own encoder, which is 11$\times$ faster.
Qualitatively, it preserves the bricks and window bars that the tiny decoders lose (\cref{fig:supp-fastvae-recon}).
In \method, FastVAE more than doubles the frame rate and lowers peak GPU memory from 42.7 to 33.6~GiB, while content persistence (89.83 vs.\ 89.76) and revisit (86.78 vs.\ 86.52) are preserved (\cref{tab:diagnostics}).

%% file: sec/supp/G_evaluation.tex
\section{Additional Details on Evaluation}
\label{sec:supp-evaluation}

\subsection{Static Environments}
\label{sec:supp-static}
\begin{figure}[t]
  \centering
  \includegraphics[width=\linewidth]{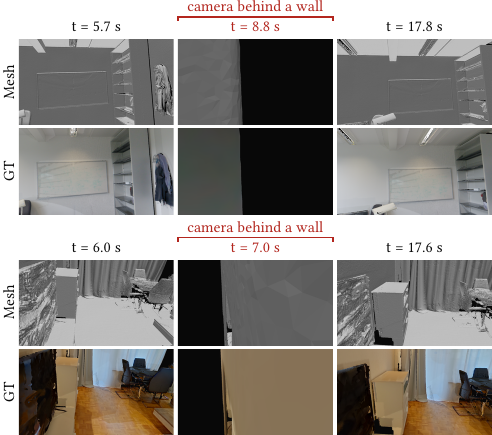}
  \caption{\textbf{Excluded trajectories.}
  We show two of the 14 ScanNet++ trajectories that we exclude, as the mesh render and the scan rendered with its colors (GT), before, while, and after the camera is behind a wall.
  Behind the wall, the view shows the back of the wall or nothing.}
  \label{fig:supp-bad-trajectory}
\end{figure}

\paragraph{Trajectories.}
On ScanNet++, we use the trajectories that Orbia~\cite{lu2026orbia} releases, one for each of 60 scenes, in which the camera leaves a view and returns to it.
In 14 of them, the camera passes through a wall, behind which both the mesh render and the ground truth show the back of the wall or nothing (\cref{fig:supp-bad-trajectory}).
We thus evaluate on the remaining 46 and report all 60 in \cref{tab:supp-static-all60}.
Including these 14 trajectories lowers every geometry-conditioned method across all metrics except input preservation while leaving SolarWM~H3 largely unchanged, yet \method still ranks first in four of the six metrics (control alignment, input preservation, content persistence, and temporal stability).
On Replica, which Orbia does not include, we generate four trajectories in each of the 18 scenes (72 in total) with Orbia's trajectory generator, and render the input frame and the ground-truth frames used for scoring from the textured mesh.
On both datasets, each trajectory has 357 frames (22\,s at 16~FPS) at $720\times1280$.

\begin{figure}[t]
  \centering
  \includegraphics[width=\linewidth]{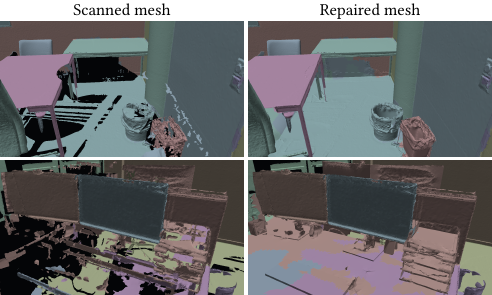}
  \caption{\textbf{Mesh repair.}
  We render two ScanNet++ scenes at the same pose from the scanned mesh (\textbf{left}) and from the mesh repaired by an LLM agent (\textbf{right}).
  The repair closes the holes in the floor and the furniture.}
  \label{fig:supp-mesh-repair}
\end{figure}

\paragraph{Mesh repair and scoring.}
On ScanNet++, we run every method that takes geometry on two versions of each mesh, the scanned one and one repaired by an LLM agent, as the scans have holes and artifacts (\cref{fig:supp-mesh-repair}).
For each trajectory and metric, we report the better of a method's two scores, for the baselines and \method alike, such that no method is penalized by the defects of a scan or by its repair.
We list the scores on each mesh separately, on the 46 trajectories, in \cref{tab:supp-static-mesh}.
The repair raises the scores of most of these methods, by up to 2.7 points for GEN3C and by at most 0.7 for \method.
Even with the scanned meshes alone, \method ranks first in control alignment, input preservation, content persistence, 3D consistency, and temporal stability.

\begin{table*}[t]
  \centering
  \caption{\textbf{ScanNet++ on each mesh.}
  We report Orbia scores~\cite{lu2026orbia} on the 46 trajectories of \cref{tab:static} with the ground-truth mesh (\textbf{left}) and the LLM-repaired mesh (\textbf{right}), whereas \cref{tab:static} keeps the better of the two for each trajectory and score.
  SolarWM takes no geometry, so its two sides are the same videos, a number in parentheses counts the trajectories on which a score is undefined and counted as 0, and $\dagger$ and $\S$ mark methods that generate at 480p and 540p.}
  \label{tab:supp-static-mesh}
  \footnotesize
  \setlength{\tabcolsep}{4.425pt}
  \newcommand{\hcell}[1]{\begin{tabular}[c]{@{}c@{}}#1\end{tabular}}
  \begin{tabular}{@{}lcccccccccccc}
    \toprule
    & \multicolumn{6}{c}{ScanNet++, ground-truth mesh (46)} & \multicolumn{6}{c}{ScanNet++, LLM-repaired mesh (46)} \\
    \cmidrule(lr){2-7}\cmidrule(lr){8-13}
    Method & \hcell{Control\\Align.} & \hcell{Visual\\Quality} & \hcell{Input\\Pres.} & \hcell{Content\\Persist.} & \hcell{3D\\Consist.} & \hcell{Temp.\\Stab.} & \hcell{Control\\Align.} & \hcell{Visual\\Quality} & \hcell{Input\\Pres.} & \hcell{Content\\Persist.} & \hcell{3D\\Consist.} & \hcell{Temp.\\Stab.} \\
    \midrule
    SolarWM H3~\cite{huang2026solarwm} & 83.01 & 71.72 & 71.46\,{\scriptsize(1)} & 80.34\,{\scriptsize(1)} & \second{74.96} & 87.22 & 83.01 & 71.72 & 71.46\,{\scriptsize(1)} & 80.34\,{\scriptsize(1)} & \second{74.96} & 87.22 \\
    \midrule
    Matrix-Game~3.5~\cite{qian2026matrixgame35} & 56.93 & 68.49 & 70.01\,{\scriptsize(1)} & 49.27\,{\scriptsize(8)} & 49.31 & 72.77 & 57.05 & 68.88 & 70.51\,{\scriptsize(1)} & 47.19\,{\scriptsize(12)} & 50.20 & 72.64 \\
    GEN3C~\cite{ren2025gen3c} & 81.54 & 66.59 & 78.93\,{\scriptsize(2)} & 83.63 & 71.35 & 85.10 & 84.21 & 66.67 & 80.25\,{\scriptsize(2)} & 83.96 & 71.20 & 86.43 \\
    Lyra~2.0$^\dagger$~\cite{shen2026lyra2} & \third{91.40} & 71.20 & 81.27\,{\scriptsize(1)} & 78.27 & 74.48 & \third{89.15} & \third{91.00} & 71.37 & 81.01\,{\scriptsize(1)} & 79.89\,{\scriptsize(1)} & 74.08 & \third{89.29} \\
    Lyra~2.0 (DMD)$^\dagger$~\cite{shen2026lyra2} & \second{91.54} & 71.28 & \third{82.04\,{\scriptsize(1)}} & \third{84.26} & \third{74.80} & \second{89.76} & \second{91.44} & 71.33 & \third{81.96\,{\scriptsize(1)}} & \third{84.26} & \second{74.96} & \second{89.66} \\
    \midrule
    WorldDirector$^\dagger$~\cite{wang2026worlddirector} & 76.52 & 66.74 & 77.14\,{\scriptsize(2)} & 71.71\,{\scriptsize(2)} & 69.08 & 81.99 & 76.15 & 66.97 & 78.80\,{\scriptsize(1)} & 68.72\,{\scriptsize(3)} & 67.26 & 81.95 \\
    Code World Model~\cite{chen2026codeworld} & 81.30 & 71.71 & 81.79\,{\scriptsize(1)} & 77.38\,{\scriptsize(2)} & 72.76 & 88.28 & 82.20 & 71.67 & 79.84\,{\scriptsize(2)} & 75.82\,{\scriptsize(2)} & 73.12 & 88.21 \\
    Code Video Model$^\S$~\cite{liang2026codevideomodel} & 64.63 & \second{73.07} & 72.32\,{\scriptsize(5)} & 67.35\,{\scriptsize(5)} & 73.71 & 84.10 & 65.19 & \second{73.27} & 71.24\,{\scriptsize(6)} & 69.30\,{\scriptsize(4)} & 73.94 & 84.49 \\
    AgentGarten$^\dagger$~\cite{chi2026agentgarten} & 87.78 & \first{73.17} & \second{86.82\,{\scriptsize(1)}} & \second{87.07} & 74.15 & 87.74 & 88.34 & \first{73.57} & \second{86.94\,{\scriptsize(1)}} & \second{89.00} & 74.60 & 88.49 \\
    \midrule
    \method & \first{94.96} & \third{72.01} & \first{88.34\,{\scriptsize(1)}} & \first{88.68} & \first{76.11} & \first{90.91} & \first{95.38} & \third{72.44} & \first{88.44\,{\scriptsize(1)}} & \first{89.35} & \first{76.36} & \first{91.28} \\
    \bottomrule
  \end{tabular}
\end{table*}

\begin{table}[t]
  \centering
  \caption{\textbf{All 60 ScanNet++ trajectories.}
  We report Orbia scores~\cite{lu2026orbia} on all 60 trajectories, including the 14 in which the camera passes through a wall, scored as in \cref{tab:static}.
  Parentheses and markers are as in \cref{tab:supp-static-mesh}.}
  \label{tab:supp-static-all60}
  \scriptsize
  \setlength{\tabcolsep}{2.76pt}
  \newcommand{\hcell}[1]{\begin{tabular}[c]{@{}c@{}}#1\end{tabular}}
  \begin{tabular}{@{}lcccccc}
    \toprule
    & \multicolumn{6}{c}{ScanNet++, all trajectories (60)} \\
    \cmidrule(lr){2-7}
    Method & \hcell{Control\\Align.} & \hcell{Visual\\Quality} & \hcell{Input\\Pres.} & \hcell{Content\\Persist.} & \hcell{3D\\Consist.} & \hcell{Temp.\\Stab.} \\
    \midrule
    SolarWM H3~\cite{huang2026solarwm} & 82.90 & \third{71.67} & 71.03\,{\tiny(1)} & 80.27\,{\tiny(1)} & \first{75.24} & 87.25 \\
    \midrule
    Matrix-Game~3.5~\cite{qian2026matrixgame35} & 58.42 & 68.97 & 71.55\,{\tiny(1)} & 48.96\,{\tiny(11)} & 49.37 & 74.14 \\
    GEN3C~\cite{ren2025gen3c} & 84.55 & 66.60 & 82.02\,{\tiny(2)} & \third{84.89} & 71.16 & 85.72 \\
    Lyra~2.0$^\dagger$~\cite{shen2026lyra2} & \third{88.48} & 71.55 & 81.20\,{\tiny(1)} & 77.22 & \second{74.41} & \second{88.33} \\
    Lyra~2.0 (DMD)$^\dagger$~\cite{shen2026lyra2} & \second{89.47} & 70.85 & 81.58\,{\tiny(1)} & 79.57 & 73.54 & \third{87.91} \\
    \midrule
    WorldDirector$^\dagger$~\cite{wang2026worlddirector} & 77.70 & 67.06 & 79.74\,{\tiny(1)} & 71.07\,{\tiny(2)} & 68.25 & 82.06 \\
    Code World Model~\cite{chen2026codeworld} & 78.77 & 71.44 & \third{82.32\,{\tiny(1)}} & 75.39\,{\tiny(2)} & 71.75 & 86.93 \\
    Code Video Model$^\S$~\cite{liang2026codevideomodel} & 66.00 & \first{73.26} & 75.06\,{\tiny(5)} & 67.14\,{\tiny(5)} & 73.37 & 84.13 \\
    AgentGarten$^\dagger$~\cite{chi2026agentgarten} & 84.41 & \second{72.12} & \second{87.36\,{\tiny(1)}} & \second{85.70\,{\tiny(1)}} & 71.86 & 86.10 \\
    \midrule
    \method & \first{92.71} & 71.33 & \first{88.81\,{\tiny(1)}} & \first{86.87} & \third{73.61} & \first{88.95} \\
    \bottomrule
  \end{tabular}
\end{table}

\paragraph{Baselines.}
We run every baseline with its released sampling steps and context window.
Lyra~2.0~\cite{shen2026lyra2} and WorldDirector~\cite{wang2026worlddirector} generate at $480\times832$ and Code Video Model~\cite{liang2026codevideomodel} at $544\times960$, the only resolutions they support, and AgentGarten~\cite{chi2026agentgarten} at $480\times832$, the resolution it is trained at.
SolarWM~\cite{huang2026solarwm} and Code World Model~\cite{chen2026codeworld} generate at $768\times1344$, and the remaining methods at $704\times1280$.
Methods that generate fixed-length blocks pad the last block with the final pose.
As Code Video Model has no input to continue a video, we generate each trajectory in 124-frame windows and give each window the last frame of the previous one as its first key frame.
We provide the ground-truth mesh to every method that takes geometry, and Matrix-Game~3.5~\cite{qian2026matrixgame35}, GEN3C~\cite{ren2025gen3c}, and Lyra~2.0 build their 3D caches from depth rendered from this mesh instead of the depth they estimate.

\paragraph{Metrics.}
We report the six scores of Orbia~\cite{lu2026orbia} with its released evaluator, each on a 0--100 scale.
Control alignment measures the camera pose error and the accuracy of the move and look directions, and visual quality combines aesthetic, image-quality, flicker, smoothness, and preference scores.
Input preservation compares later views of the surfaces visible in the input with ground truth, and content persistence (generated-content persistence in Orbia) compares each revisit of a region absent from the input with its first generated view.
3D consistency (3D self-consistency in Orbia) measures the agreement of depth, features, and scale across views, and how well a Gaussian scene predicted from the generated frames renders held-out views.
Temporal stability tracks quality, control, and local geometry over 5\,s windows.
As its long-horizon part requires minute-long trajectories, which Orbia does not provide on ScanNet++, we report its short-horizon part.

\subsection{Dynamic Environments}
\label{sec:supp-dynamic}
\paragraph{Datasets.}
We evaluate on five of the articulated HSSD homes~\cite{khanna2024hssd} released with PARTNR~\cite{chang2025partnr}, which we render with Habitat~\cite{szot2021habitat2}.
Their drawers and doors open and close, and their textured meshes provide ground truth.
We also evaluate on \ourworlds, eight worlds that we build with a coding agent (two alpine villages, a driving district, three harbors, a parcel depot, and a woodland campsite).
As these worlds have no textures, we render their geometry with flat shading and turn the first frame of each trajectory into a realistic image with Gemini~3 Pro Image.
This image is the only appearance that a method receives.

\paragraph{Trajectories.}
We render three types of single-camera trajectories in both datasets, in which the camera reopens drawers and doors, returns to rooms it has seen, or revisits objects that may have moved (\cref{tab:supp-dynamic-trajectories}).
Every revisit replays the camera poses of the first visit, such that the ground-truth frames of the two visits differ only where the scene has changed.
On HSSD, a revisit comes 19 to 94\,s after the first visit.
We render all trajectories at 16~FPS and $720\times1280$, with a camera that moves at up to 0.75\,m/s and turns at up to 40$^\circ$/s.

\begin{table}[t]
  \centering
  \caption{\textbf{Trajectories in dynamic environments.}
  We list the number of trajectories of each type on HSSD and in \ourworlds.
  As room-revisit trajectories also reopen drawers and doors, object permanence is scored on 24 trajectories on HSSD and 7 in \ourworlds.
  Trajectories last 34 to 113\,s on HSSD and 44 to 108\,s in \ourworlds, and every revisit comes at least 19\,s after the first visit.}
  \label{tab:supp-dynamic-trajectories}
  \footnotesize
  \setlength{\tabcolsep}{3pt}
  \begin{tabular*}{\linewidth}{@{\extracolsep{\fill}}lccp{0.38\linewidth}@{}}
    \toprule
    Type & HSSD & \ourworlds & What happens \\
    \midrule
    Reopen & 13 & 1 & The camera opens drawers or doors, leaves, and reopens them. \\
    Room revisit & 11 & 6 & The camera also glances at the room and repeats the glances later. \\
    Moved object & 21 & 45 & An object moves in view, moves unseen, or stays in place. \\
    \midrule
    Total & 45 & 52 & \\
    \bottomrule
  \end{tabular*}
\end{table}

\paragraph{Inputs.}
Every method receives the first frame and the camera path of a trajectory.
All methods that take geometry also receive the same mesh as \method, with its moving parts posed at every frame (or at each chunk update for 3D-cache methods), so they receive the updated scene geometry directly.
We run every baseline with the settings of \cref{sec:supp-static}.

\paragraph{Metrics.}
We measure memory by comparing the frames that a method generates at each revisit with its own frames at the same poses on the first visit, inside ground-truth masks.
\emph{Revisit} is the cosine similarity of DINOv2 patch features~\cite{oquab2024dinov2} over the static parts of a room view.
\emph{Object permanence} averages this similarity inside a reopened drawer or door with a rating by Gemini~3.7 Flash~\cite{google2026gemini37flash} of whether the contents are the same.
For \emph{moved objects}, we compare each revisited spot against two ground-truth renders at the same pose, one with the object and one without it, and compute the fraction of DINOv2 patches inside the object mask that are closer to the expected state.
We average this score with a Gemini rating of whether it is the same object.
Gemini rates each pair of crops three times from 0 to 10 with the questions of \cref{tab:supp-vlm-questions}, and we scale all scores to 0--100.
As a check, the ground-truth renders reach at least 98.8 on all three scores.

\begin{table}[t]
  \centering
  \caption{\textbf{Rated questions.} We list the questions that Gemini~3.7 Flash answers with a rating from 0 to 10, given two crops of a method's own video side by side.
  Agent persistence (\cref{sec:supp-demos}) is zero unless its first question is rated at least 5.}
  \label{tab:supp-vlm-questions}
  \footnotesize
  \begin{tabular}{@{}p{0.30\linewidth}p{0.64\linewidth}@{}}
    \toprule
    Score & Question \\
    \midrule
    Object permanence & Overall, would a viewer believe the second and later openings show the very same contents as the first? \\
    Moved objects & Is the object at the centre of the right panel the very same physical object as the one at the centre of the left panel (same kind of object, shape, proportions, colours, materials and markings)? \\
    Agent persistence & Is there a person (a human character; not an object, a box, a grey block or empty scenery) at the centre of both panels? \\
    Agent persistence & Is the person at the centre of the right panel the very same character as the one at the centre of the left panel (same build, hair, clothing, colours and accessories)? \\
    \bottomrule
  \end{tabular}
\end{table}

\subsection{Multi-Agent Environments}
\label{sec:supp-demos}
\paragraph{Episodes.}
In \ourworlds, we render 30 episodes with two agents and 23 with four, in which each agent later sees what another agent saw first.
Each agent is followed by its own third-person camera, and the first frame of each view is made as for a single camera (\cref{sec:supp-dynamic}).
In 22 of the two-agent episodes and all 23 four-agent episodes, each agent first inspects a place of its own, the agents meet, each visits the place of the next agent along the same camera path, they meet again at the same spots, and each returns to its own place.
Seven of these two-agent episodes and two of the four-agent episodes also include a drawer that one agent opens and another reopens.
The remaining eight two-agent episodes focus on moved objects across agents, where an object either moves unseen or stays in place while the two agents take turns inspecting the scene.
The second look at a place comes 20 to 64\,s after the first, and the two meetings are 24 to 54\,s apart.
Episodes last 63 to 128\,s, and we render every view at 16~FPS and $720\times1280$.

\paragraph{Metrics.}
\emph{Cross-agent consistency} measures whether a place looks the same to an agent as it did to the agent that saw it first.
We compute the revisit, object permanence, and moved objects scores of \cref{sec:supp-dynamic} with the first visit taken from the video of another agent and all agents masked, and average them (the first two with four agents, whose episodes have no moved object).
\emph{Agent persistence} measures whether an agent looks the same to another agent at both of their meetings.
We average the DINOv2 similarity inside the box of that agent with a rating by Gemini of whether both meetings show the same character (\cref{tab:supp-vlm-questions}).
The score is zero unless both meetings show a person, as a method that draws no agent would otherwise be scored on the scenery behind the box.

\paragraph{Baselines.}
No baseline can share its memory between views.
Each thus generates the view of every agent as a separate single-camera trajectory (\cref{sec:supp-dynamic}), whereas \method generates all views of an episode together from one surface memory (\cref{sec:supp-memory}).
Each baseline receives the agents in the form that its inputs take.
AgentGarten, Code World Model, and Code Video Model receive a human-shaped figure in place of the box that \method receives (\cref{sec:supp-rendering}), WorldDirector receives a tracked box with a prompt that names a person, and the others receive only the prompt and the first frame.

%% file: sec/supp/H_limitations.tex
\section{Limitations}
\label{sec:supp-limitations}
Although a coarse mesh suffices, \method assumes that it is accurate, and errors in the mesh carry over into the generated video.
It already tolerates the holes and artifacts of real scans, as repairing the ScanNet++ meshes raises its scores by at most 0.7 points (\cref{tab:supp-static-mesh}).
Handling noisier geometry reconstructed from images would be a valuable next step.
With multiple agents, \method also requires the first frame of every view to be well aligned with the mesh.
Frontier image models such as Gemini~3 Pro Image, which we use (\cref{sec:supp-dynamic}), often produce such frames, whereas obtaining them from older open-source models remains difficult.
As the renderer receives each agent only as a box, it has to guess the appearance of a side of an agent that no view has shown.
We currently keep the agent's appearance consistent across views with a few rules at inference (\cref{sec:supp-memory}), which explicit identity controls could replace.
Finally, \method renders non-human characters less reliably and less faithfully than humans.
Only 34 of the 401 characters and less than 5\% of the training videos are non-human (\cref{sec:supp-data}), and we introduced them only in the final 2{,}000 iterations of the 513-frame bidirectional stage (\cref{tab:supp-training}) and the subsequent causal stages, exclusively in single-agent episodes.
However, \method already generalizes to many of them, which makes training on more such data a promising direction.

%% file: sec/supp/I_societal_impact.tex
\section{Societal Impact}
\label{sec:supp-impact}
\method renders coarse, untextured meshes into photorealistic video in real time, which benefits applications such as gaming, embodied AI, and closed-loop simulation~\cite{turki2026simuli,basant2026omnidreams}.
Since an explicit world specifies geometry and state, a user can inspect and edit what happens in a generated video, and \method determines only how it looks.
However, as \method renders realistic video of places and people from a single initial frame, it raises concerns about potential misuse for generating deceptive visual media (e.g., presenting a generated video as real footage).
Appropriate safeguards such as watermarking generated content~\cite{wen2023treering} should be considered for deployment.
Finally, \method builds on a pretrained video model~\cite{nvidia2026cosmos3} and is fine-tuned only on synthetic data~\cite{karp2026isekai3d}, and thus inherits the biases of both.

%% file: sec/supp/J_qualitative.tex
\section{Additional Qualitative Results}
\label{sec:supp-qualitative}
\begin{figure*}[p]
  \centering
  \includegraphics[page=1,width=\linewidth]{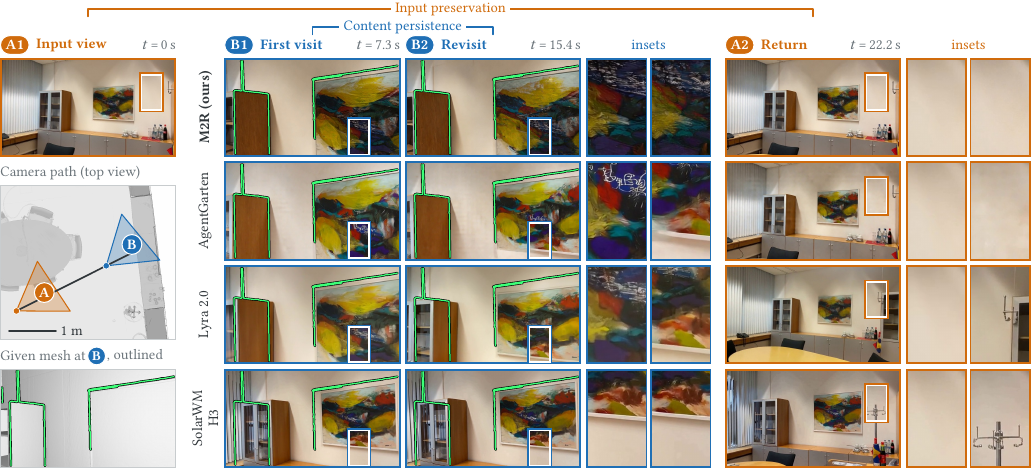}\\[6pt]  %
  \includegraphics[page=2,width=\linewidth]{figures/gallery_scannetpp_v1_supp_compact.pdf}  %
  \caption{\textbf{ScanNet++.} We show two rollouts in the layout of \cref{fig:static}, in an office with a painting (\textbf{top}) and in a kitchen that the camera revisits from a different angle (\textbf{bottom}).
  AgentGarten repaints the painting between the two visits (\textbf{top insets}), and SolarWM~H3 stays far back from it.
  AgentGarten's revisit shows another part of the kitchen, and GEN3C does not return to the input view (\textbf{bottom}).
  \method follows the camera and keeps both views in both rollouts.}
  \label{fig:supp-gallery-scannetpp-1}
\end{figure*}

\begin{figure*}[p]
  \centering
  \includegraphics[page=1,width=\linewidth]{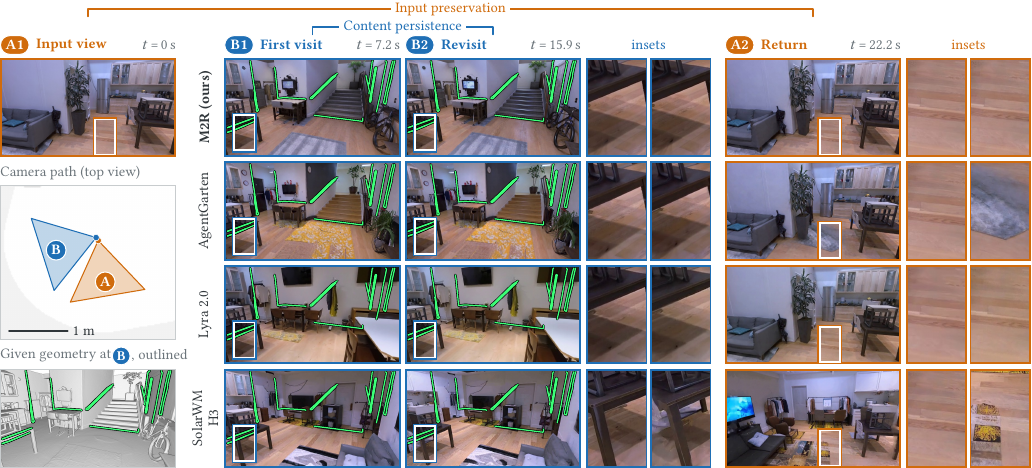}\\[6pt]  %
  \includegraphics[page=2,width=\linewidth]{figures/gallery_replica_v1_supp_compact.pdf}  %
  \caption{\textbf{Replica.} We show two rollouts in the layout of \cref{fig:static}, in an apartment with stairs (\textbf{top}) and along a wall with a painting (\textbf{bottom}).
  On these short loops, most methods keep the revisited view, and they differ mainly in how they follow the given geometry and return to the input view.
  SolarWM~H3 and Lyra~2.0 draw rooms that differ from the given geometry, and a rug appears in AgentGarten's return (\textbf{top insets}).
  AgentGarten draws a dark screen where the other methods draw a painting (\textbf{bottom}).
  \method follows the geometry and keeps both views.}
  \label{fig:supp-gallery-replica-1}
\end{figure*}

\begin{figure*}[p]
  \centering
  \includegraphics[page=1,width=0.94\linewidth]{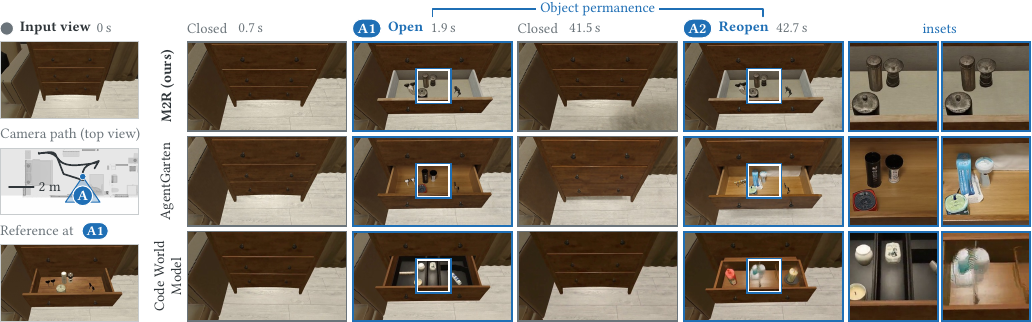}\\[4pt]  %
  \includegraphics[page=2,width=0.94\linewidth]{figures/gallery_hssd_v1_supp_compact.pdf}  %
  \caption{\textbf{HSSD.} We show a drawer that is reopened after 41\,s (\textbf{top}) and a vase that is moved from a counter (A) into a fridge (B) while the camera looks away (\textbf{bottom}), in the layout of \cref{fig:dynamic}.
  Code World Model and AgentGarten change the contents of the drawer and keep the vase on the counter or draw a darker one, while \method preserves both.
  The frame at the bottom left is the reference render at the first visit.}
  \label{fig:supp-gallery-hssd-1}
  \par\vspace{10pt}
  \includegraphics[page=1,width=0.94\linewidth]{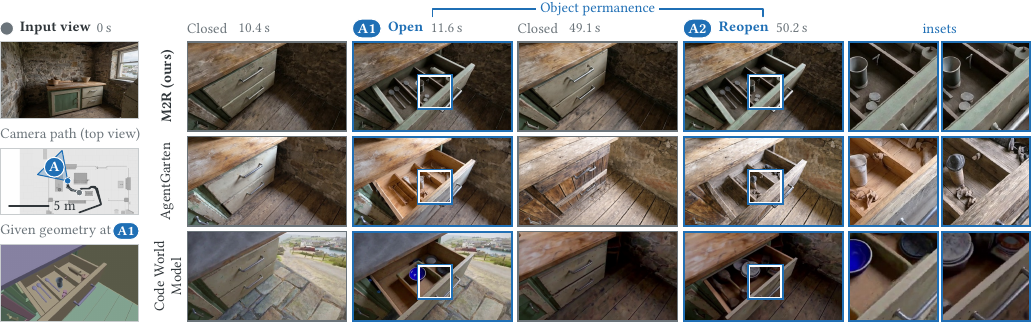}\\[4pt]  %
  \includegraphics[page=2,width=0.94\linewidth]{figures/gallery_our_worlds_v1_supp_compact.pdf}  %
  \caption{\textbf{\ourworlds.} We show a bakery drawer that is reopened after 39\,s (\textbf{top}) and a room that is revisited from two views (\textbf{bottom}).
  Code World Model moves from a quay into a room and changes the furniture, and AgentGarten changes the drawer, its contents, and the view through the window, while \method preserves all of them.}
  \label{fig:supp-gallery-our-worlds-1}
\end{figure*}

\begin{figure*}[p]
  \centering
  \includegraphics[page=1,width=0.93\linewidth]{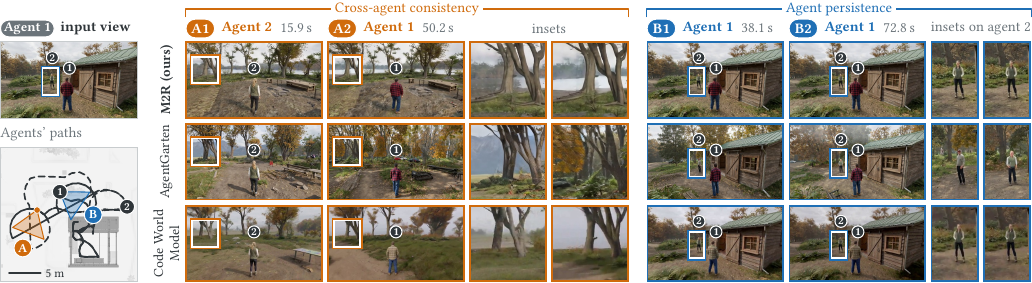}\\[4pt]  %
  \includegraphics[page=2,width=0.93\linewidth]{figures/gallery_multiagent_v1_supp.pdf}  %
  \caption{\textbf{Multiple agents.} We show two rollouts in the layout of \cref{fig:multiagent}, in a woodland campsite (\textbf{top}) and on a street (\textbf{bottom}), with an input view (the box marks the other agent) and both agents' paths (agent 2's dashed).
  AgentGarten and Code World Model show each agent a different place, and AgentGarten also draws another person or a signpost at the first meeting.
  Code World Model keeps the other agent, and \method keeps both the place and the other agent.}
  \label{fig:supp-gallery-multiagent-1}
  \par\vspace{10pt}
  \includegraphics[page=3,width=0.93\linewidth]{figures/gallery_multiagent_v1_supp.pdf}\\[4pt]  %
  \includegraphics[page=4,width=0.93\linewidth]{figures/gallery_multiagent_v1_supp.pdf}  %
  \caption{\textbf{Multiple agents with a dynamic object.} In a harbor workshop, we show cross-agent consistency for a place (\textbf{A}) and for a tool drawer that agent 1 opens and agent 2 reopens 38\,s later (\textbf{B}), and agent persistence from both cameras at two meetings 34\,s apart (\textbf{C}, \textbf{D}).
  AgentGarten and Code World Model show agent 2 another place, draw the drawer without its tools, and change or lose the other agent in at least one view, while \method keeps the place, the tools, and both agents.
  The left column shows both input views, the agents' paths, and the given geometry at the first meeting, in which the agents are boxes.}
  \label{fig:supp-gallery-multiagent-2}
\end{figure*}

We show additional rollouts in static, dynamic, and multi-agent environments in \cref{fig:supp-gallery-scannetpp-1,fig:supp-gallery-replica-1,fig:supp-gallery-hssd-1,fig:supp-gallery-our-worlds-1,fig:supp-gallery-multiagent-1,fig:supp-gallery-multiagent-2}.

